\documentclass[sigconf]{acmart}

\AtBeginDocument{%
\providecommand\BibTeX{{%
		\normalfont B\kern-0.5em{\scshape i\kern-0.25em b}\kern-0.8em\TeX}}}

\copyrightyear{2025}
\acmYear{2025}
\setcopyright{acmlicensed}
\acmConference[MM '25] {Proceedings of the 33rd ACM International Conference on Multimedia}{October 27--31, 2025}{Dublin, Ireland.}
\acmBooktitle{Proceedings of the 33rd ACM International Conference on Multimedia (MM '25), October 27--31, 2025, Dublin, Ireland}
\acmISBN{979-8-4007-2035-2/2025/10}
\acmDOI{10.1145/3746027.3755608}

\acmSubmissionID{4742}

\usepackage{algorithm}
\usepackage{algpseudocode}
\usepackage{amsmath}

\let\digamma\relax

\usepackage{amssymb}
\usepackage{subcaption}
\usepackage{tabularx}
\usepackage{color}
\usepackage{multirow}
\usepackage{threeparttable}
\usepackage[american]{babel}
\usepackage{microtype}
\usepackage{bbm}

\newcommand{\x}{{\boldsymbol{x}}}

\newcommand{\z}{{\boldsymbol{z}}}

\newcommand{\I}{{\boldsymbol{I}}}

\newcommand{\The}{{\boldsymbol{\theta}}}
\newcommand{\eps}{{\boldsymbol{\epsilon}}}

\renewcommand{\geq}{\geqslant}
\renewcommand{\leq}{\leqslant}

\begin{document}
\title{Frequency Regulation for Exposure Bias Mitigation in Diffusion Models}

\author{Meng Yu}
\orcid{0009-0001-5121-5816}
\email{yum21@lzu.edu.cn}
\affiliation{%
	\institution{School of Information Science and Engineering, \\
		Lanzhou University}
	\city{Lanzhou}
	\country{China}
}
\author{Kun Zhan}
\orcid{0000-0002-8000-5682}
\authornote{Corresponding author.}
\email{kzhan@lzu.edu.cn}	
\affiliation{%
	\institution{School of Information Science and Engineering, \\
		Lanzhou University}
	\city{Lanzhou}
	\country{China}
}

\renewcommand{\shortauthors}{Meng Yu \& Kun Zhan.}

\begin{abstract}
	Diffusion models exhibit impressive generative capabilities but are significantly impacted by exposure bias. In this paper, we make a key observation: the energy of predicted noisy samples in the reverse process continuously declines compared to perturbed samples in the forward process. Building on this, we identify two important findings: 1) The reduction in energy follows distinct patterns in the low-frequency and high-frequency subbands; 2) The subband energy of reverse-process reconstructed samples is consistently lower than that of forward-process ones, and both are lower than the original data samples. Based on the first finding, we introduce a dynamic frequency regulation mechanism utilizing wavelet transforms, which separately adjusts the low- and high-frequency subbands. Leveraging the second insight, we derive the rigorous mathematical form of exposure bias. It is worth noting that, our method is training-free and plug-and-play, significantly improving the generative quality of various diffusion models and  frameworks with negligible computational cost. The source code is available at \url{https://github.com/kunzhan/wpp}.
\end{abstract}


\maketitle

\section{Introduction}\label{sec:intro}
In recent years, Diffusion Probabilistic Models (DPMs)~\cite{sohl2015deep,ho2020denoising} have made remarkable progress in image generation~\cite{dhariwal2021diffusion,rombach2022high}. ADM~\cite{dhariwal2021diffusion} made the generation quality of DPMs surpass Generative Adversarial Networks (GANs)~\cite{goodfellow2014generative} by introducing classifier guidance. EDM~\cite{Karras2022edm} notably promotes development by clarifying the design space of DPMs. Additionally, IDDPM~\cite{nichol2021improved}, DDIM~\cite{songdenoising}, Analytic-DPM~\cite{baoanalytic}, EA-DPM~\cite{bao2022estimating}, PFGM++~\cite{xu2023pfgm++}, and AMED~\cite{zhou2024fast} also promote the improvement of DPMs from different aspects. However, these models still suffer from exposure bias~\cite{ning2023input}, the mismatch between the forward and reverse process in DPMs. Due to the prediction error~\cite{kim2023refining} of the network and the
discretization error~\cite{zhang2023fast} of the numerical solver, the reverse trajectory
of DPMs tends to deviate from the ideal path. Meanwhile, the bias will gradually accumulate during sampling, ultimately affecting the generation quality.

\begin{figure}[!t]
	\centering
	\begin{subfigure}[b]{0.48\linewidth}
		\centering
		\includegraphics[width=\linewidth]{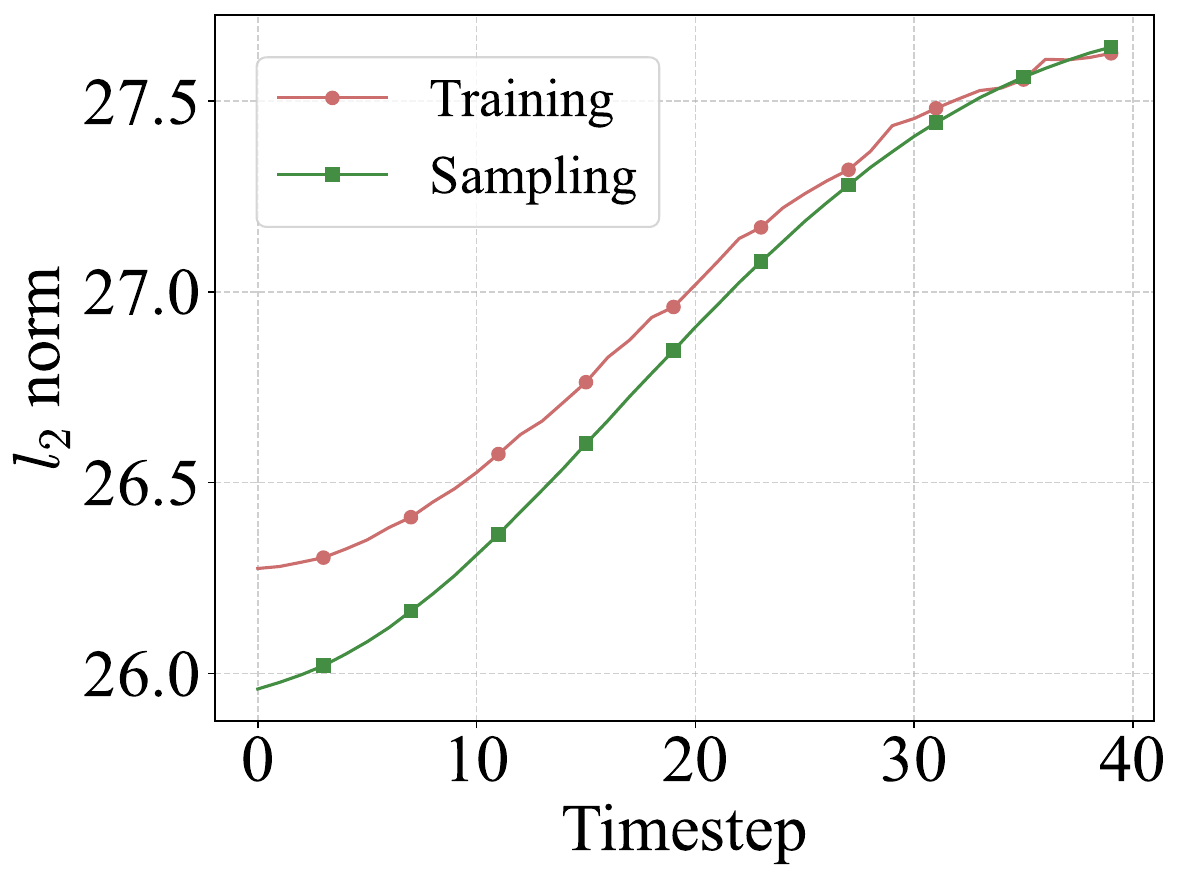}
		\caption{The low-frequency energy.}
		\label{fig4a:xll}
	\end{subfigure}%
	\begin{subfigure}[b]{0.48\linewidth}
		\centering
		\includegraphics[width=\linewidth]{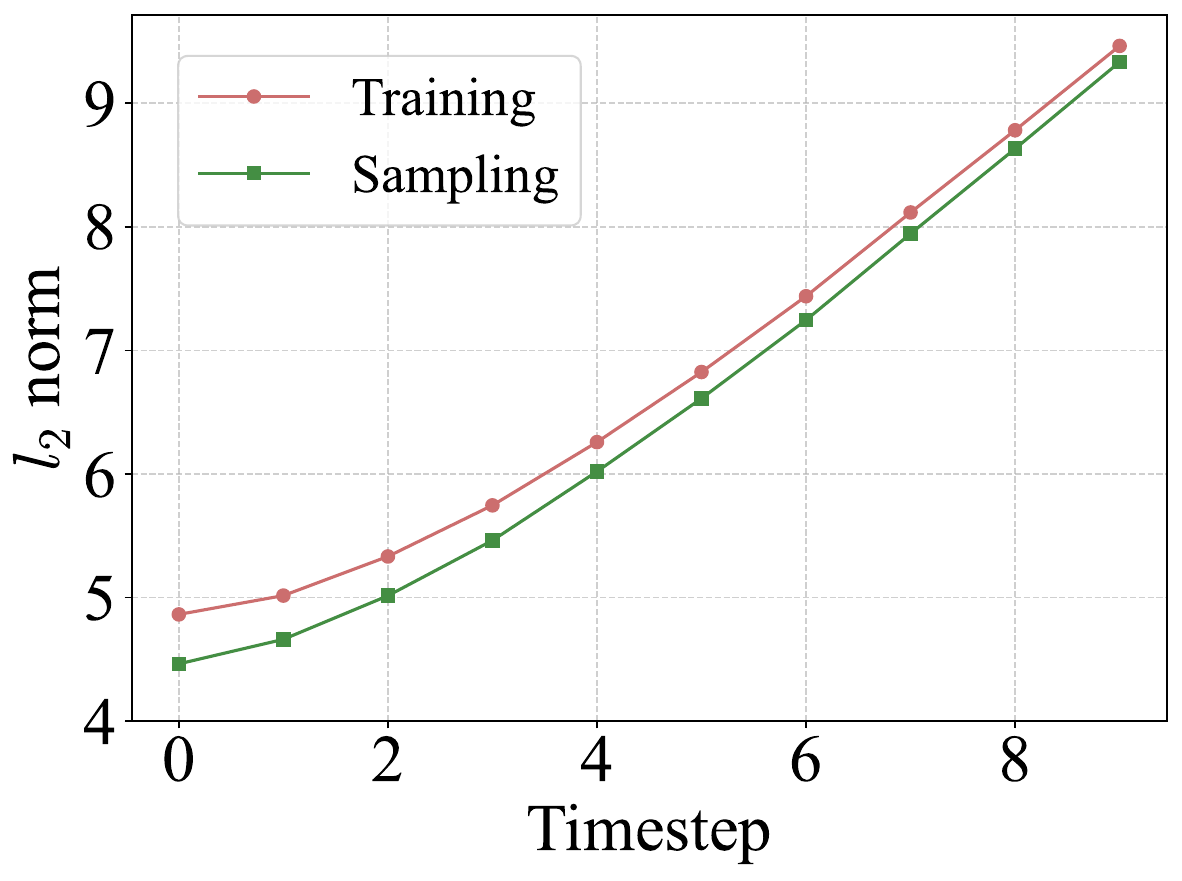}
		\caption{The high-frequency energy.}
		\label{fig4b:xlh}
	\end{subfigure}
	\caption{The subband energy of the noisy sample in training and sampling, with denoising from right to left.}
	\label{fig1:norm_show}
\end{figure}

A variety of methods have been proposed to mitigate exposure bias. These approaches include diverse training strategies~\cite{ning2023input, li2023error, ren2024multi} and sampling techniques~\cite{li2024alleviating, zhang2025antiexposure, yao2025manifold}. However, all these methods are confined to operating in the spatial domain, neglecting the analysis and mitigation of exposure bias in the wavelet domain. We observed during the reverse process, the energy of the predicted noisy sample is always lower than that of the forward perturbed sample. Interestingly, this reduction pattern exhibits distinctly different modes in different frequency subbands. By applying Discrete Wavelet Transform (DWT), we analyze the evolution of exposure bias within the wavelet domain and uncover the first key findings: low-frequency subband energy reduction persists throughout sampling, with high-frequency energy primarily diminishing in the later reverse stage, as shown in Fig.~\ref{fig1:norm_show}. Applying the same method, we analyze the reconstructed sample, which directly predicts the original sample given the noisy sample. Thus, we have the second key observation: the subband energy of reverse-process reconstructed samples is consistently lower than that of forward-process ones, and both are lower than the original data samples.

Based on the first finding, we propose a simple yet effective frequency information adjustment mechanism. The predicted noisy sample is decomposed into high- and low-frequency subbands via DWT. Then, the low-frequency subband is amplified with a dedicated weight throughout sampling, while the high-frequency subbands are amplified in the late sampling using the high-frequency weight. Finally, the adjusted subbands are combined and reconstructed into the predicted noisy sample in the spatial space via inverse Discrete Wavelet Transform.

Building on the second finding, we derive the mathematical analytical form of exposure bias. Previous work has severely lacked exploration into the analytical form of exposure bias. In particular, their modeling of exposure bias relies heavily on strong assumptions~\cite{ningelucidating,li2024alleviating}. Leveraging the discovery of the energy law of reconstructed samples, we conduct more accurate modeling and derivation, providing a clearer understanding of exposure bias.

Additionally, we observe amplifying the low-frequency subband towards the end of the sampling process may conflict with the denoising principles of DPMs, potentially introducing undesirable effects. Specifically, DPMs aim to finely restore image details in the later stages of sampling. As shown in Fig.~\ref{fig2:evolution_dpm}, high-frequency information begins to change significantly only at the final stages. If a relatively large weight is assigned to the low-frequency subband during this phase, it hinders the refinement process. To address this issue, we explore several dynamic weighting strategies, where the amplifying effect of the low-frequency subband decreases as sampling progresses, while that of high-frequency components gradually increases. 

In summary, our contributions are:

\begin{figure}[!t] 
	\centering
	\begin{subfigure}[b]{0.09\textwidth}
		\includegraphics[width=\linewidth]{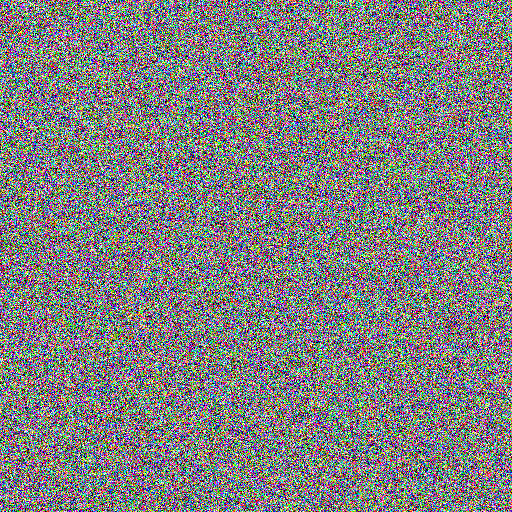}
	\end{subfigure}
	\begin{subfigure}[b]{0.09\textwidth}
		\includegraphics[width=\linewidth]{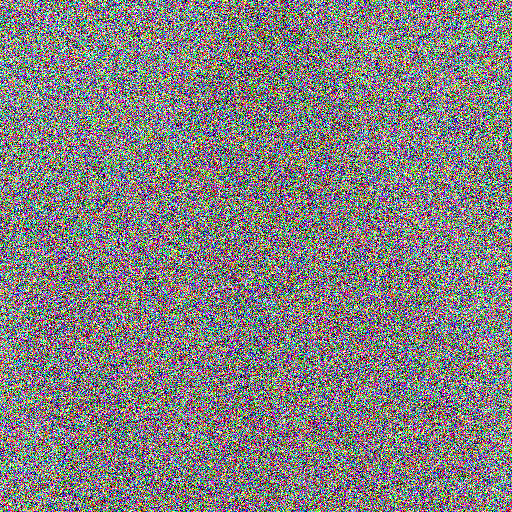}
	\end{subfigure}
	\begin{subfigure}[b]{0.09\textwidth}
		\includegraphics[width=\linewidth]{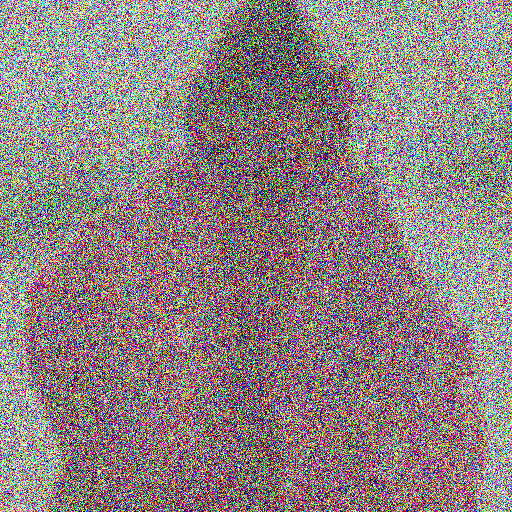}
	\end{subfigure}
	\begin{subfigure}[b]{0.09\textwidth}
		\includegraphics[width=\linewidth]{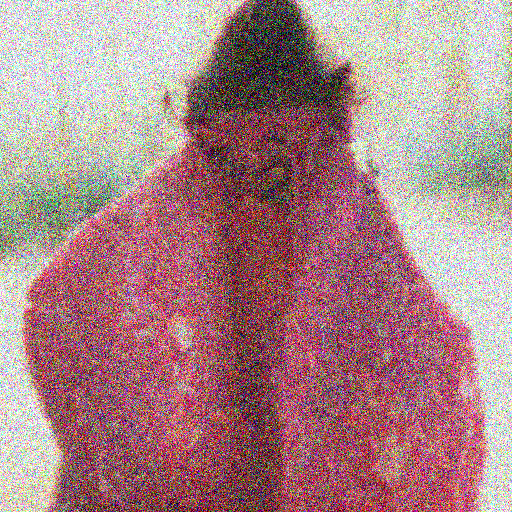}
	\end{subfigure}
	\begin{subfigure}[b]{0.09\textwidth}
		\includegraphics[width=\linewidth]{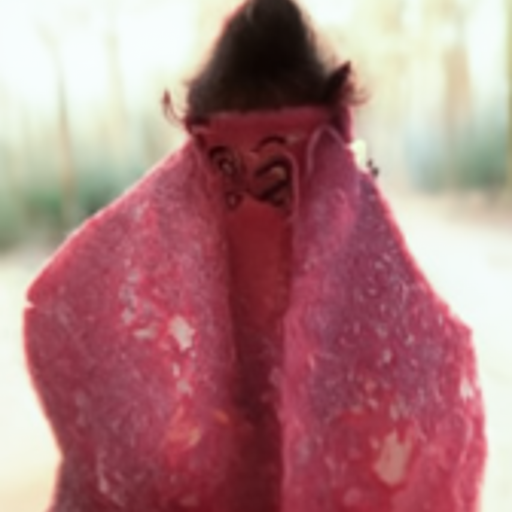}
	\end{subfigure}

	\begin{subfigure}[b]{0.09\textwidth}
		\includegraphics[width=\linewidth]{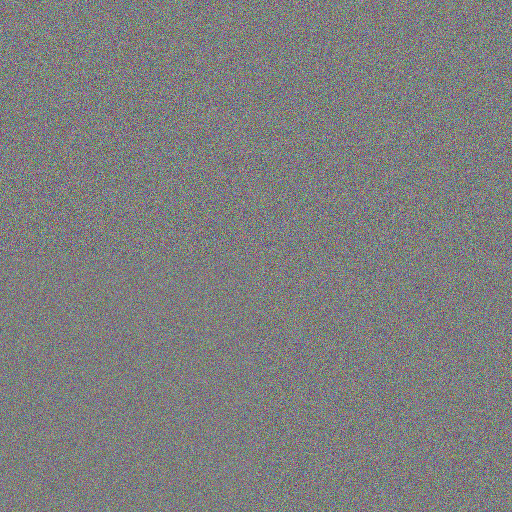}
	\end{subfigure}
	\begin{subfigure}[b]{0.09\textwidth}
		\includegraphics[width=\linewidth]{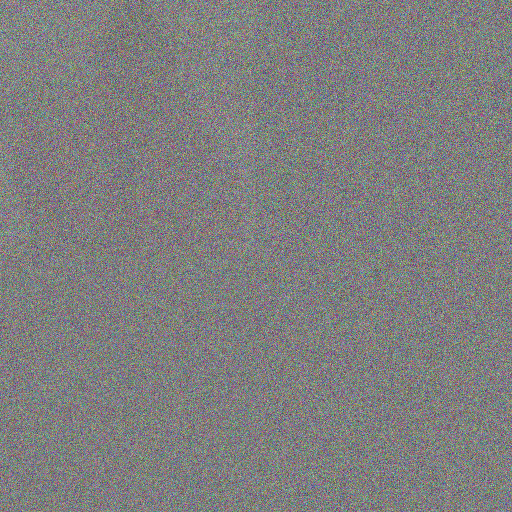}
	\end{subfigure}
	\begin{subfigure}[b]{0.09\textwidth}
		\includegraphics[width=\linewidth]{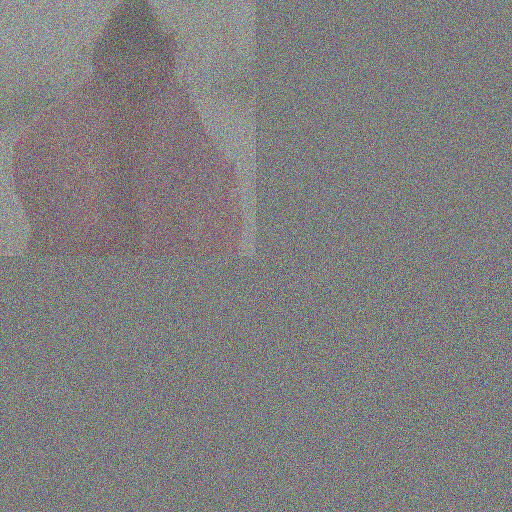}
	\end{subfigure}
	\begin{subfigure}[b]{0.09\textwidth}
		\includegraphics[width=\linewidth]{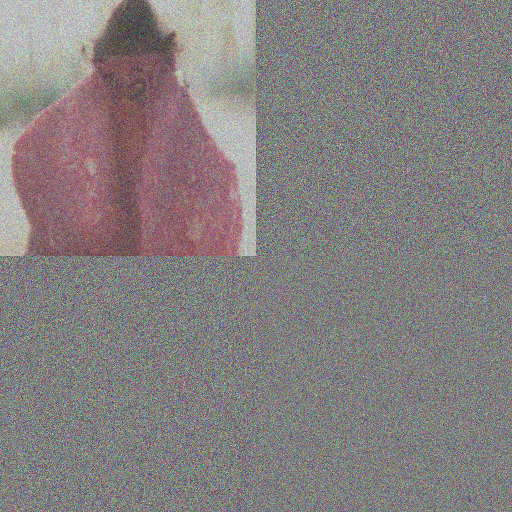}
	\end{subfigure}
	\begin{subfigure}[b]{0.09\textwidth}
		\includegraphics[width=\linewidth]{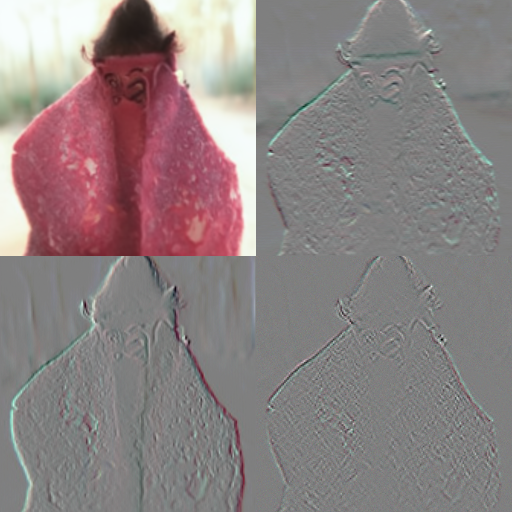}
	\end{subfigure}
	
	\caption{The evolution of the predicted noisy sample $\hat{\x}_t$ (the first row) and the four subbands (the second row) of $\hat{\x}_t$ in the wavelet domain during sampling. Applying discrete wavelet transform on $\hat{\x}_t$ yields four distinct frequency components: $\hat{\x}_t^{ll}$ ($\nwarrow$), $\hat{\x}_t^{lh}$ ($\nearrow$), $\hat{\x}_t^{hl}$ ($\swarrow$), and $\hat{\x}_t^{hh}$ ($\searrow$).}
	\label{fig2:evolution_dpm}
\end{figure}

\begin{itemize}
	\item We are, to the best of our knowledge, the first to analyze and solve the exposure bias of DPMs from the perspective of the wavelet domain. Furthermore, based on findings in the wavelet domain, we derive for the first time the rigorous mathematical form of exposure bias.
	\item We propose a dynamic frequency regulation mechanism based on the wavelet transform to mitigate exposure bias. The mechanism is train-free and plug-and-play, improving the generation quality of various DPMs.
	\item Our method can achieve high-quality performance metrics with negligible computational cost, outperforming all current improved models for exposure bias.
\end{itemize}

\section{Related Work}\label{sec2:related}
This section first reviews the development of DPMs, then presents some work on improving DPMs using wavelet transforms, and finally introduces existing methods for mitigating exposure bias.

The theoretical foundation of DPMs was established by DPM~\cite{sohl2015deep}, with substantial advancements later achieved by DDPM~\cite{ho2020denoising}. By introducing classifier guidance, ADM~\cite{dhariwal2021diffusion} made the generation performance of DPMs surpass GAN~\cite{goodfellow2014generative} 
for the first time. SDE~\cite{song2021scorebased} proposes a unified DPM framework by means of stochastic differential equations, while DDIM~\cite{songdenoising} accelerates sampling by skipping steps. Analytic-DPM~\cite{baoanalytic} theoretically derived the optimal variance in the reverse denoising process, while EDM~\cite{Karras2022edm} clarified the design space of DPMs, which further enhances the generation quality of DPMs. In addition, ODE-based DPMs~\cite{lu2022dpmsolver,zhou2024fast,zhao2024unipc,zheng2023fast,dockhorn2022genie}, DPMs incorporating distillation techniques~\cite{salimans2022progressive,liu2023instaflow,meng2023distillation,luhman2021knowledge,zheng2023fast}, and consistency models~\cite{song2023consistency,song2024improved,lu2025simplifying} are widely developed, which significantly speeds up sampling and improves the generation quality.

Wavelet decomposition~\cite{graps1995introduction,mallat1989theory} has been widely applied in the field of GANs~\cite{gal2021swagan,yang2022wavegan,wang2022fregan,zhang2022styleswin}, improving the quality of various generation tasks. In recent years, some works have started to combine the wavelet transform with DPMs. WACM ~\cite{li2022wavelet} utilizes the wavelet spectrum to assist image colorization. WSGM~\cite{guth2022wavelet} factorizes the data distribution into the product of conditional probabilities of wavelet coefficients at various scales to accelerate the generation process. WaveDiff~\cite{phung2023wavelet} processes images in the wavelet domain to further enhance the efficiency and quality of generation.

The exposure bias in DPMs was first systematically analyzed by ADM-IP~\cite{ning2023input}, which simulated the bias by re-perturbing the training sample. Subsequently, EP-DDPM~\cite{li2023error} regard the cumulative error as the regularization term to retrain the model for alleviating exposure bias, while MDSS~\cite{ren2024multi} employed the multi-step denoising scheduled sampling strategy to mitigate this issue. It should be noted these three methods require retraining the model. Conversely, TS-DPM~\cite{li2024alleviating} put forward the time-shift sampler and ADM-ES applied the noise scaling technique, which both can mitigate exposure bias without retraining models. Additionally, AE-DPM~\cite{zhang2025antiexposure} alleviates the bias through prompt learning, and MCDO~\cite{yao2025manifold} mitigates it by means of manifold constraint. S++~\cite{yubias} proposes a score difference correction mechanism to reduce exposure bias.

Unfortunately, all previous studies on exposure bias have been confined to the spatial domain, also called the pixel domain. In contrast, we analyze this issue from the perspective of the wavelet domain and address it using frequency approaches.
\section{Preliminaries}\label{sec:PRELIMINARIES}
In this section, we present the essential background necessary to understand our problem and approach. We first review Diffusion Probabilistic Models (DPMs)~\cite{ho2020denoising}, then discuss exposure bias~\cite{ning2023input}, and finally introduce the wavelet transform.
\subsection{Diffusion Model}
Diffusion Probabilistic Model (DPM) usually consists of a forward noising process and a reverse denoising process, both of which are defined as Markov processes. For a original data distribution $q(\x_0)$ and a noise schedule $\beta_t$, the forward process is
\begin{equation}
	q\left(\x_{1:T}|\x_0\right)=\prod_{t=1}^Tq\left(\x_t|\x_{t-1}\right),
	\label{eq1:forward_markov}
\end{equation}
where $q\left(\x_t|\x_{t-1}\right)=\mathcal{N}(\x_t;\sqrt{1-\beta_t}\x_{t-1},\beta_t\I)$. Specifically, by leveraging the properties of the Gaussian distribution, the perturbed distribution can be rewritten as the conditional distribution $q\left(\x_t|\x_0\right)$, which is expressed by the formula
\begin{equation}
	\x_t=\sqrt{\bar{\alpha}_t}\x_0+\sqrt{1-\bar{\alpha}_t}\eps_t,
	\label{eq2:forward_onestep}
\end{equation}
where ${\alpha}_t=1-\beta_t$, $\bar{\alpha}_t = \prod_{i=1}^t\alpha_i$, and $\eps_t\sim \mathcal{N}(\boldsymbol{0},\I)$. Based on Bayes’ theorem, the posterior conditional probability is obtained by 
\begin{equation}
	q(\x_{t-1}|\x_t,\x_0)=\mathcal{N}(\tilde{\mu}_t(\x_t,\x_0),\tilde{\beta}_t\I),
	\label{eq3:real_posterior}
\end{equation}
where $\tilde{\mu}_{t}=\frac{\sqrt{\overline{\alpha}_{t-1}}\beta_{t}}{1-\overline{\alpha}_{t}}\x_{0}+\frac{\sqrt{\alpha_{t}}(1-\overline{\alpha}_{t-1})}{1-\overline{\alpha}_{t}}\x_{t}$ and $\tilde{\beta}_t=\frac{1-\overline{\alpha}_{t-1}}{1-\overline{\alpha}_t}\beta_t$. We apply a neural network $\eps_{\The}(\x_t,t)$ to approximate $q(\x_{t-1}|\x_t,\x_0)$, and naturally we want to minimize $D_{\rm KL}(q(\x_{t-1}|\x_t,\x_0)|| p_{\The}(\x_{t-1}|\x_t)))$. Through a series of reparameterizations and derivations, we obtain
\begin{equation}
	\begin{aligned}
		\mu_{\The}(\x_t,t) &= \frac{\sqrt{\overline{\alpha}_{t-1}}\beta_{t}}{1-\overline{\alpha}_{t}} \x^0_\The(\x_t,t)+ \frac{\sqrt{\alpha_{t}}(1-\overline{\alpha}_{t-1})}{1-\overline{\alpha}_{t}}\x_{t}\\
		&= \frac{1}{\sqrt{\alpha_t}} \left(\x_t - \frac{1 - \alpha_t}{\sqrt{1 - \bar{\alpha}_t}} \eps_\The(\x_t, t) \right),
	\end{aligned}
\end{equation}
where $\eps_\The(\cdot)$ is the noise
prediction network and $\x^0_\The(\x_t,t)$ represent the reconstruction model which predicts $\x_0$ given $\x_t$:
\begin{equation}
	\x^0_\The(\x_t,t)=\frac{x_t-\sqrt{\bar{\alpha}_t}\eps_\The(\x_t,t)}{\sqrt{\bar{\alpha}_t}}.
	\label{eq5:reconstruction}
\end{equation}
So the final loss function is given by
\begin{equation}
	\mathcal{L}_{\rm simple}=\mathbb{E}_{t,\x_0,\eps_t\sim\mathcal{N}(\mathbf{0},\I)}[\|\eps_{\The}(\x_t,t)-\eps_t\|_2^2].
	\label{eq4:loss_function}
\end{equation} Once $\eps_{\The}(\cdot)$ has converged, we can perform reverse sampling starting from a standard Gaussian distribution by using $p_{{\The}}(\x_{t-1}|\x_t)$.  
\subsection{Exposure Bias}
\begin{figure}[t]
	\centering
	\includegraphics[width=0.8\linewidth]{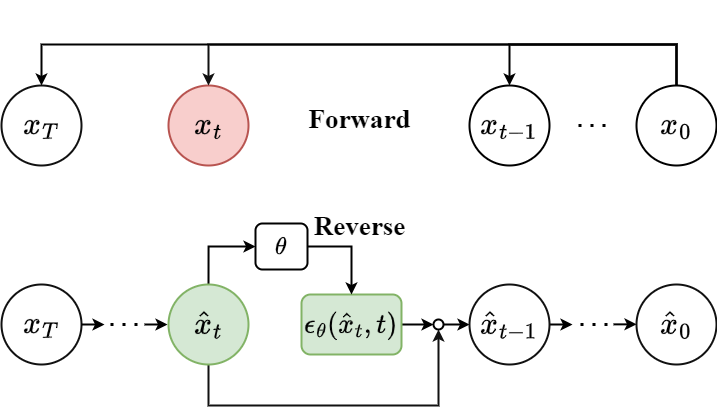} 
	\caption{The schematic diagram of exposure bias in DPM.}
	\label{fig1:bias_question}
\end{figure}

The exposure bias is the mismatch between the forward and reverse processes in DPMs. Specifically, for the pre-trained noise prediction network $\eps_\The(\cdot)$, the network input comes from the perturbed noisy sample $\x_t$ based on the ground truth during training, while during sampling the network input comes from the predicted noisy sample $\hat{\x}_t$. Due to the prediction error of the neural network and the discretization error of the numerical solver, the sampling trajectory of DPM tends to deviate from the ideal path, which leads to the bias between the predicted noisy sample $\hat{\x}_t$ and the perturbed noisy sample $\x_t$. This exposure bias between noisy samples leads to a bias in the network output $\eps_\The\left(\hat{\x}_{t},t\right)$ and $\eps_\The\left(\x_{t},t\right)$, which exacerbates the exposure bias in the next time step. Therefore, the exposure bias gradually accumulates with the sampling iteration, and ultimately affects the generation quality, as shown in Fig.~\ref{fig1:bias_question}. Thus, due to exposure bias, the actual sampling formula should be
\begin{equation}
	\hat{\x}_{t-1} = \frac{1}{\sqrt{\alpha_t}} \left(\hat{\x}_t - \frac{1 - \alpha_t}{\sqrt{1 - \bar{\alpha}_t}} \eps_\The(\hat{\x}_t, t) \right) + \sigma_t \z.
	\label{eq6:actual_inference}
\end{equation}

Specially, both $\hat{\x}_T$ and $\x_T$ obey the standard Gaussian distribution. However, due to the prediction error between $\eps_\The(\x_{T}, T)$ and the ideal noise, the sampling operation in Eq.~\eqref{eq6:actual_inference} leads to the deviation between $\hat{\x}_{T-1}$ and $\x_{T-1}$, which further exacerbates the bias between $\eps_\The(\hat{\x}_{T-1}, T-1)$ and $\eps_\The(\x_{T-1}, T-1)$. Thus, as sampling continues, the bias of the predicted noisy sample and the bias of the network prediction influence each other and gradually accumulate, causing the exposure bias between $\hat{\x}_t$ and $\x_t$. 

\subsection{Wavelet Transform}
We introduce Discrete Wavelet Transform (DWT), which decomposes an image $\x \in \mathbb{R}^{H\times W}$ into four wavelet subbands, namely $\x^{ll}, \x^{lh}, \x^{hl}$, and $\x^{hh}$. Conversely, these wavelet subbands can be reconstructed into an image by using the inverse Discrete Wavelet Transform (iDWT). The two processes are expressed as:
\begin{equation}
	\begin{cases}
		\{\x^f | f \in\{ ll, lh, hl, hh\}\} = \textrm{DWT}(\x) \\
		\x = \textrm{iDWT}(\x^f|f \in\{ ll, lh, hl, hh\})
	\end{cases}
	\label{eq7:dwt_idwt}
\end{equation}
where the resolution of subbands is $ \mathbb{R}^{H/2\times W/2}$. $\x^{ll}$ represents the low-frequency component of the image $\x$, reflecting the basic structure of the image, such as the human face shape and the shape of birds. $\x^{lh}, \x^{hl}$, and $\x^{hh}$ represent the high-frequency components in different directions of the image, reflecting the detailed information of the image, such as human wrinkles and bird feathers. For simplicity, we choose the classic Haar wavelet in experiments.
\section{Method}\label{sec:method}
We first analyze exposure bias in the wavelet domain using Discrete Wavelet Transform (DWT). The key finding reveals that the low-frequency subband energy of predicted samples in the reverse process is consistently lower than that of perturbed samples in the forward process throughout the entire sampling process, while the high-frequency subband energy decreases only at the end of sampling. Then, we similarly use DWT to analyze the energy evolution patterns of reconstructed samples in both forward and reverse processes, based on which a reasonable hypothesis is proposed to derive the analytical form of exposure bias. Finally, a simple yet effective frequency-energy adjustment mechanism is proposed to mitigate exposure bias, which works by adjusting the frequency-energy distribution at each time step. Additionally, considering the rule that the reverse process of DPMs first reconstructs low-frequency information and then focuses on recovering high-frequency, a novel dynamic weighting strategy is proposed to further precisely scale the energy of frequency subbands.
\subsection{Energy Reduction}
\label{sec:4.1}
\begin{figure}[!t]
	\centering
	\begin{subfigure}[b]{0.48\linewidth}
		\centering
		\includegraphics[width=\linewidth]{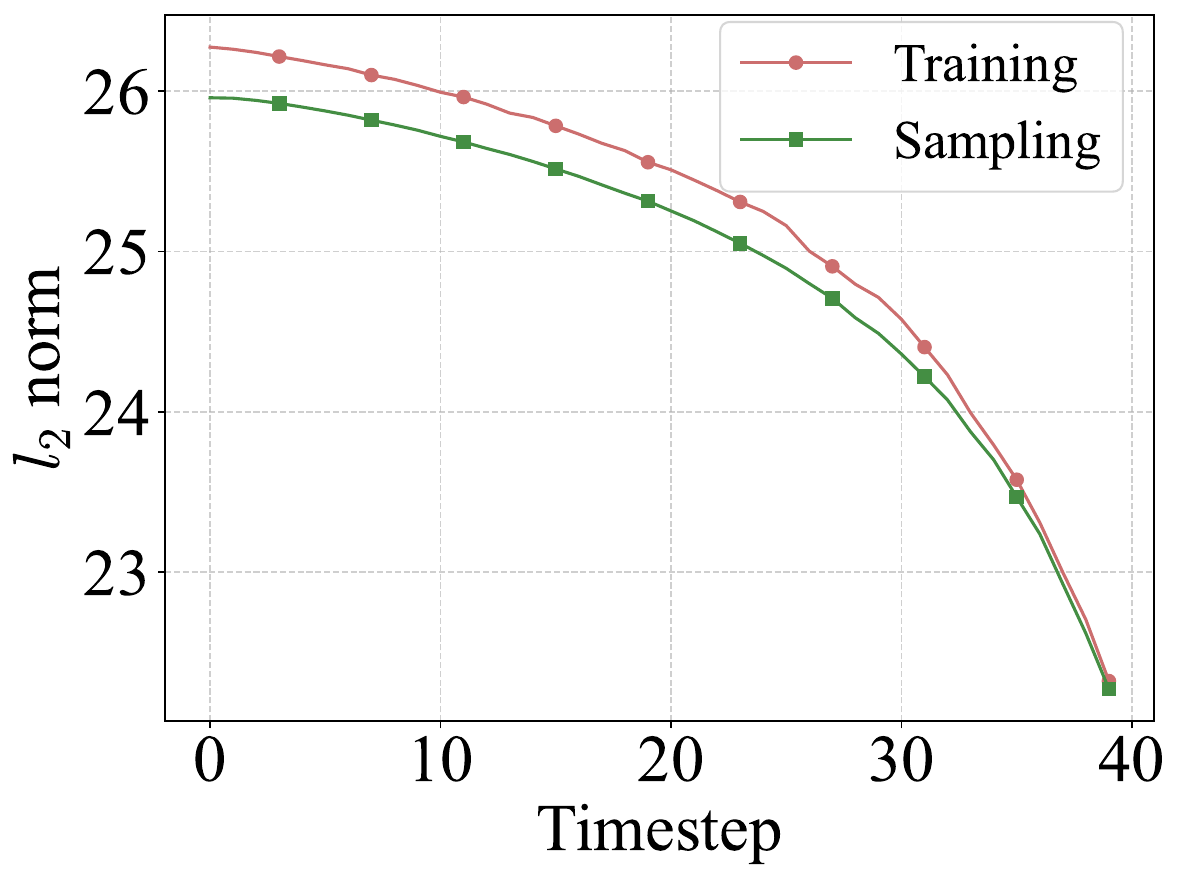}
		\caption{The $ll$ subband of $\x^0_\The({\x}_t,t)$}
		\label{fig4c:x0_ll}
	\end{subfigure}%
	\begin{subfigure}[b]{0.48\linewidth}
		\centering
		\includegraphics[width=\linewidth]{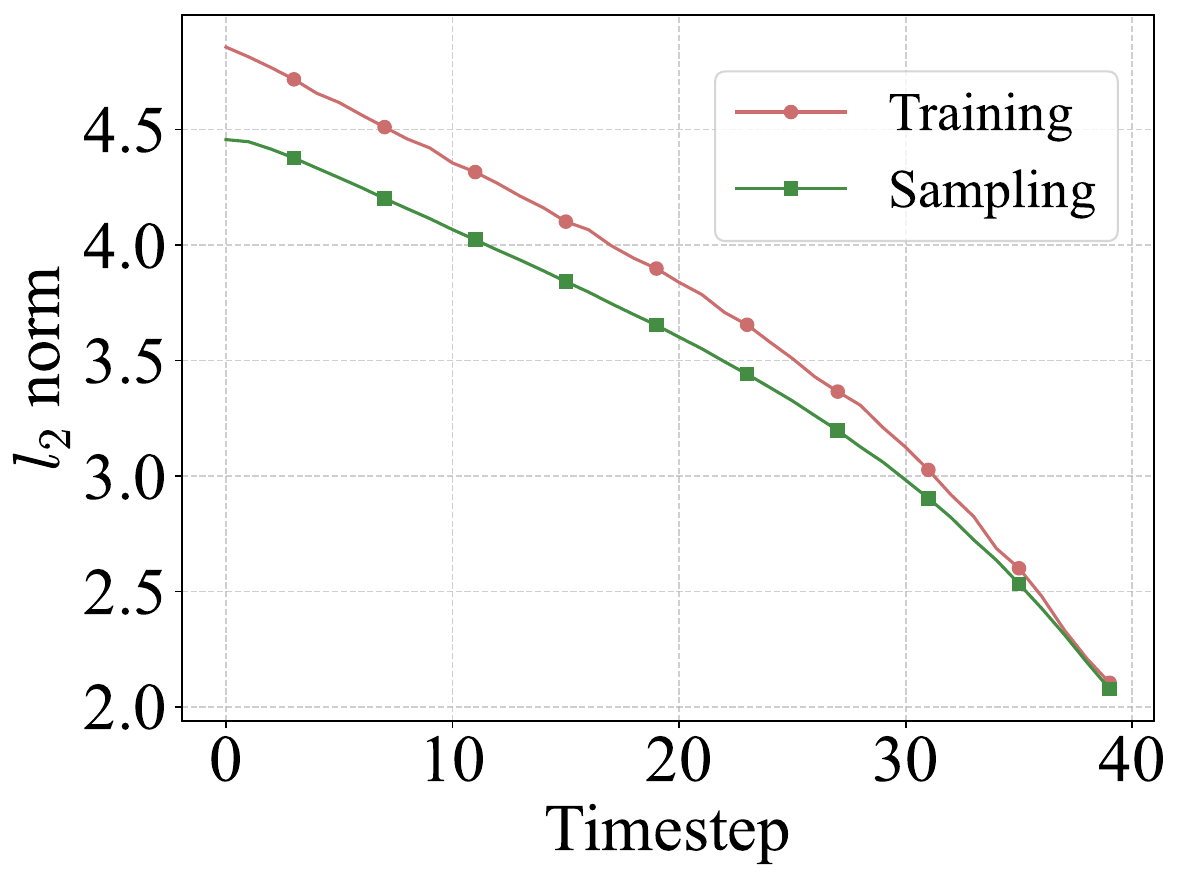}
		\caption{The $lh$ subband of $\x^0_\The({\x}_t,t)$}
		\label{fig4d:x0_lh}
	\end{subfigure}%
	\caption{The subband energy of $\x^0_\The({\x}_t,t)$ in training and sampling, with denoising from right to left.}
	\label{fig:x0}
\end{figure}

Here, we analyze exposure bias in the wavelet domain. Firstly, we need to simulate exposure bias accurately. During sampling, it is easy to obtain the predicted noisy $\hat{\x}_t$, but acquiring the ideal perturbed noisy $\x_t$ corresponding to $\hat{\x}_t$ is difficult. To solve this problem, we rely on the original data and the deterministic solver to simulate the exposure bias. Specifically, for the given original data $\x_0$, we first perform forward perturbation on $\x_0$ to obtain a series of perturbed noisy $\{\x_1, \x_2,\ldots, \x_{s+1}\}$. Then, we take $\x_{s+1}$ as the actual sampling starting point, match the corresponding time step $s+1$ and coefficients, and carry out the DDIM sampling~\cite{songdenoising} 
\begin{equation}
	\hat{\x}_{s-1}=\sqrt{\bar{\alpha}_{s-1}}\left(\frac{x_s-\sqrt{\bar{\alpha}_s}\eps_\The(\x_s,s)}{\sqrt{\bar{\alpha}_s}}\right)+\sqrt{1-\bar{\alpha}_{s-1}}\eps_\The(\x_s,s).
	\label{eq8:forward_onestep_s}
\end{equation}As a result, a series of predicted noisy sample $\{\hat{\x}_{s-1},\ldots,\hat{\x}_2, \hat{\x}_1, \hat{\x}_0\}$ is obtained. As long as we ensure that $\x_s$ still maintains a certain level of low-frequency information of the original data $\x_0$, we are able to make the reverse predicted noisy sample $\hat{\x}_t$ similar to the forward perturbed noisy sample $\x_t$. 

Secondly, we perform Discrete Wavelet Transform (DWT) on the perturbed and predicted noisy sample at different time steps to get their corresponding frequency components. Then, we calculate respectively the mean squared norm of the subbands of these two batches of noisy samples at different time steps. Fig.~\ref{fig4a:xll} clearly shows that as the sampling progresses, the energy of the low frequency subbands of $\hat{\x}_t$ gradually becomes lower than that of $\x_t$, which indicates there is a phenomenon of reduction of low-frequency energy during sampling. Fig.~\ref{fig4b:xlh} shows only at the end of sampling does the high-frequency subband energy of $\hat{\x}_t$ become significantly lower than that of ${\x}_t$ (during the early and middle stages of sampling, the two are comparable). Intuitively, the diffusion model starts to reconstruct low-frequency information in the early stage of sampling, so the deviation of the low-frequency subband will accumulate and propagate throughout the entire sampling process. In contrast, high-frequency information is only restored emphatically at the end stage, so the deviation of the high-frequency subband is only manifested at the end stage of sampling.

The above experiments indicate there is a reduction of low frequency energy throughout the sampling process and a reduction of high-frequency energy at the end of sampling, which is the manifestation of exposure bias in the wavelet domain.

\begin{table}[t]
	\centering
	\caption{Mean and variance of $q(\x_t|\x_0)$ and $p_{\The}(\hat{\x}_t|\x_{t + 1})$}
	\begin{tabular}{lcc}
		\toprule
		& Mean & Variance \\
		\midrule
		$q(\x_t|\x_0)$ & $\sqrt{\bar{\alpha}_t}\x_0$ & $(1 - \bar{\alpha}_t)\boldsymbol{I}$ \\
		$p_{\The}(\hat{\x}_t|\x_{t + 1})$ & $\gamma_t \sqrt{\bar{\alpha}_t}\x_0$ & $\left(1 - \bar{\alpha}_t + \left(\frac{\sqrt{\bar{\alpha}_t}\beta_{t + 1}}{1 - \bar{\alpha}_{t + 1}}\phi_{t + 1}\right)^2\right)\boldsymbol{I}$ \\
		\bottomrule
	\end{tabular}
	\label{tab1:Mean_Var}
\end{table}

\subsection{Theoretical Perspective}
Then, we use the same method to explore the bias pattern of reconstructed samples and derive the analytical form of exposure bias. Previous work~\cite{ningelucidating,li2024alleviating} rely heavily on a strong assumption
\begin{equation}
	\x^0_\The(\x_t,t) = \x_0 + \phi_t \eps_t,
\end{equation}
where $\x^0_\The(\x_t,t)$ represent
the reconstruction model predicting $\x_0$ given $\x_t$, shown in Eq.~\eqref{eq5:reconstruction}, and $\phi_t$ is a scalar coefficient related to time steps. However, we will overthrow this assumption.

To explore the modeling rule of $\x^0_\The(\x_t,t)$, we used Eq.~\eqref{eq5:reconstruction} to obtain $\x^0_\The(\x_t,t)$ and $\x^0_\The(\hat{\x}_t,t)$ corresponding to $\x_t$ and $\hat{\x}_t$ based on the experiment in \S\ref{sec:4.1}, and performed the same wavelet decomposition and norm calculation Figs.~\ref{fig4c:x0_ll} and \ref{fig4d:x0_lh} clearly show that neither $\x_t$ nor $\hat{\x}_t$ can fully reconstruct $\x_0$ at any time step, and $\x^0_\The(\hat{\x}_t,t)$ reconstructed by the predicted noisy sample is smaller than $\x^0_\The(\x_t,t)$ reconstructed by the perturb noisy sample regardless of the frequency component and time step. Based on this key observation, we make a more scientific assumption about $\x^0_\The({\x}_t,t)$.

\textbf{Assumption 1} \textit{Whether in the forward process or the reverse process, $\x^0_\The({\x}_t,t)$ is assumed to be represented in terms of $\x_0$} as
\begin{equation}
	\x^0_\The(\hat{\x}_t,t) = \gamma_t \x_0 + \phi_t \eps_{t},
	\label{eq11:assumption}
\end{equation}
\textit{where $\eps_{t} \sim \mathcal{N}(\boldsymbol{0},\I)$,  $0<\gamma_t\leq 1$, $\phi_t<M$, and $M$ is an upper bound.}
Based on this assumption, we re-parameterize the conditional probability in the sampling process: 
\begin{equation}
	\hat{\x}_{t-1} = \frac{\sqrt{\overline{\alpha}_{t-1}}\beta_{t}}{1-\overline{\alpha}_{t}} \x^0_\The(\x_t,t)+ \frac{\sqrt{\alpha_{t}}(1-\overline{\alpha}_{t-1})}{1-\overline{\alpha}_{t}}\x_{t}+\sqrt{\tilde{\beta_t}}\eps_1.
	\label{eq12.}
\end{equation}
Through plugging Eqs.~\eqref{eq11:assumption} and \eqref{eq2:forward_onestep} into Eq.~\eqref{eq12.}, we obtain the analytical form of $\hat{\x}_t$:
\begin{equation}
	\hat{\x}_{t-1}=\gamma_{t-1}\sqrt{\bar{\alpha}_{t - 1}}\x_0+\sqrt{1-\bar{\alpha}_{t-1}+
		\left(\frac{\sqrt{\bar{\alpha}_{t-1}}\beta_{t}}{1-\bar{\alpha}_{t}}\phi_{t}\right)^{2}}{\eps}_2,
	\label{eq13.bias_reverse}
\end{equation}
where $\eps_1,\eps_2 \sim\mathcal{N}(\mathbf{0},\I)$. The proof of ~\ref{eq13.bias_reverse} is provided in the appendix. Table~\ref{tab1:Mean_Var}  clearly shows that the predicted distribution always has a larger variance and a smaller mean than the perturbation distribution, indicating that the prediction noisy sample reduces the energy about the target data $\x_0$ during sampling. Naturally, as the diffusion model focuses on recovering low-frequency energy in early stages, predictions mainly lose low-frequency components then; in late stages, when the model reconstructs high-frequency details, predictions primarily lose high-frequency components.
\subsection{Frequency Regulation}
Based on previous key findings, there are different patterns of reduction in the frequency subband energies of the predicted samples. Naturally, we hope to enhance low-frequency subband energy throughout the sampling process and high-frequency subband energy at its end. Thus, we propose a simple and effective mechanism for wavelet domain component regulation based on Discrete Wavelet Transform (DWT), as shown in Fig.~\ref{fig2-1:method}. Specifically, for the predicted noisy sample $\hat{\x}_t$ at each step in sampling process, we use the  DWT to decompose it into four distinct frequency components: $\hat{\x}_t^{ll}$, $\hat{\x}_t^{lh}$, $\hat{\x}_t^{hl}$, and $\hat{\x}_t^{hh}$. For simplicity, we perform the same operation for three high frequency components. In order to resist the gradual loss of low-frequency information during the entire sampling process, we reassign a global adjustment factor $w^l_t$ greater than $1$ to the low-frequency component to enhance the proportion of the low-frequency component. In order to cope with the reduction of high-frequency energy at the end of sampling, we also assign a coefficient $w^h_t$ greater than $1$ to the high-frequency component. However, this value is related to the sampling process: 
\begin{figure}[!t]
	\centering
	\includegraphics[width=0.98\linewidth]{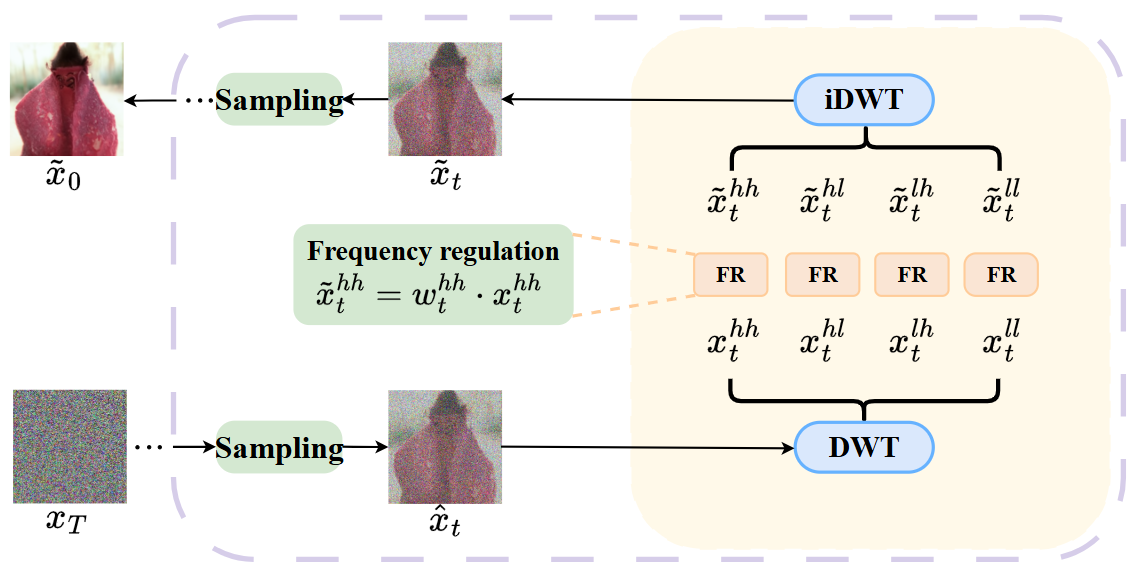} 
	\caption{
		The W++ framework. At each sampling timestep \(t\), the predicted noisy sample \(\hat{\x}_t\) is decomposed into four frequency subbands using DWT. Each subband is then assigned a weight coefficient, which is determined based on both the timestep and the type of subband. This weighting is performed via a dot product operation. After the adjustment of each subband, all the modified subbands are combined and restored to the noisy sample \(\tilde{\x}_t\) through iDWT, completing the process of frequency regulation.
	}
	\label{fig2-1:method}
\end{figure}
\begin{equation}
	w_t^h = w_h\cdot \mathbbm{1}\{t\leq t_{\text{mid}}\} 
	\label{eq9:forward_onestep_s}
\end{equation}
Then, we perform inverse Discrete Wavelet Transform (iDWT) on the adjusted frequency subbands to get the new adjusted sample:
\begin{equation}
	\tilde{\x}_t = \textrm{iDWT}(w_t^f\hat{\x}_t^f|f\in\{ll,lh,hl,hh\})
\end{equation}
Fig.~\ref{fig5:weight_show} clearly shows that with the increase in the weight coefficient $w^l_t$ of the low-frequency component within a certain range, the basic structure and color distribution of the generated noisy sample are gradually clearer and prominent, which significantly enhances the visual quality of the image. On the other hand, the change of the high-frequency component weight $w^h_t$ has a limited impact on the generated image, but the appropriate variation brings about the optimization of texture details to prevent the image from being too smooth due to the loss of high-frequency energy.

\begin{figure}[t] 
	\centering
	\begin{subfigure}[b]{0.11\textwidth} 
		\includegraphics[width=\linewidth]{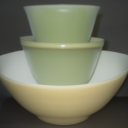}
		\caption{$w_t^l=$1.0}
	\end{subfigure}
	\hfill
	\begin{subfigure}[b]{0.11\textwidth}
		\includegraphics[width=\linewidth]{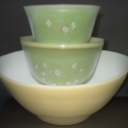}
		\caption{$w_t^l=$1.002}
	\end{subfigure}
	\hfill
	\begin{subfigure}[b]{0.11\textwidth}
		\includegraphics[width=\linewidth]{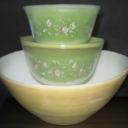}
		\caption{$w_t^l=$1.004}
	\end{subfigure}
	\hfill
	\begin{subfigure}[b]{0.11\textwidth}
		\includegraphics[width=\linewidth]{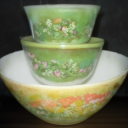}
		\caption{$w_t^l=$1.006}
	\end{subfigure}
	\begin{subfigure}[b]{0.11\textwidth}
		\includegraphics[width=\linewidth]{Figures/4.low_high/base.png}
		\caption{$w_t^h=$1.0}
	\end{subfigure}
	\hfill
	\begin{subfigure}[b]{0.11\textwidth}
		\includegraphics[width=\linewidth]{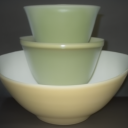}
		\caption{$w_t^h=$1.002}
	\end{subfigure}
	\hfill
	\begin{subfigure}[b]{0.11\textwidth}
		\includegraphics[width=\linewidth]{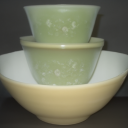}
		\caption{$w_t^h=$1.004}
	\end{subfigure}
	\hfill
	\begin{subfigure}[b]{0.11\textwidth}
		\includegraphics[width=\linewidth]{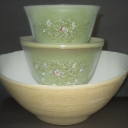}
		\caption{$w_t^h=$1.006}
	\end{subfigure}
	\caption{Effect of frequency component adjustment weight $w_l$ and $w_h$ on generation quality.}
	\label{fig5:weight_show}
\end{figure}

\subsection{Weighting Scheme}
We find that an excessively large weight for the low-frequency or high-frequency components would cause image distortion, leading to a significant reduction in the quality of the generated sample. We speculate that this is because the relatively large weight of frequency components conflicts with the denoising rules of DPM. Therefore, we need to explore more advanced weight strategies to conform to the denoising rules of DPMs.

\textbf{Low frequency weight.} DPMs focus on finely restoring the details of the image in the end stage of the sampling process, as shown in Fig~\ref{fig2:evolution_dpm}. Thus, if we still assign a relatively high weight to the low-frequency component of the image at this stage, it may impede the refinement process of the image. This conflict arises from the relative interaction of low- and high-frequency components, and  over-amplifying the former will weaken the latter. To avoid this kind of conflict, we have studied several dynamic weighting strategies that make it decrease as sampling proceeds.

The first is a simple turn-off strategy where the low-frequency component is no longer enhanced after a critical time step:
\begin{equation}
	w_t^l = w_l\cdot \mathbbm{1}\{t\geq t_{\text{mid}}\}, 
	\label{eq:forward_onestep_s}
\end{equation}
where $t_{\text{mid}}$ is a pre-defined time step, which is used to determine when to stop enhancing the low-frequency component. Another one is to utilize the variance in the reverse process:
\begin{equation}
	w_t^l = 1+w_l \cdot \sigma_t,
	\label{eq17:dwt_idwt}
\end{equation}
where $\sigma_t$ is either fixed or learned, which is determined by the baseline model. The strategy is based on the variance reflecting the denoising process. Specifically, a larger variance means more reduction to the low-frequency component of $\x_t$ in the forward process. Thus, in the reverse process, DPMs should focus more on restoring the low-frequency component at the current timestep. As the variance decreases, the low-frequency component stabilizes, allowing DPMs to focus on restoring the high-frequency components.

\textbf{High frequency weight.} DPMs focus on restoring the low-frequency part in the early stage of sampling as shown in Fig.~\ref{fig2:evolution_dpm}. In particular, high-frequency components of the image at this stage are mostly Gaussian noise rather than accurate detailed information because DPMs do not restore useful high-frequency information at this stage. Therefore, enhancing high-frequency components at this stage will amplify the noise, which obviously has negative effects and hinders DPMs from restoring the low-frequency information. Thus, we propose a turn-off strategy similar to $w_t^l$:
\begin{equation}
	w_t^h = w_h\cdot \mathbbm{1}\{t\leq t_{\text{mid}}\}, \tag{\ref{eq9:forward_onestep_s}}
\end{equation}
where $t_{\text{mid}}$ is consistent with that in Eq.~\eqref{eq:forward_onestep_s}. Therefore, $t_{\text{mid}}$ represents the critical timestep at which the enhancement of the low-frequency component stops and the enhancement of high-frequency components begins. We can also design a weight strategy that increases with the time step based on the variance:
\begin{equation}
	w_t^h = 1+w_h \cdot (1 - \sigma_t),
	\label{eq9:dwt_idwt}
\end{equation}
corresponding to Eq.~\eqref{eq17:dwt_idwt} of the low-frequency component. However, we find that the performance of Eq.~\eqref{eq9:dwt_idwt} consistently underperforms Eq.~\eqref{eq9:forward_onestep_s}. It may stem from the fact that high-frequency components in the early stages of sampling predominantly represent noise, which should not be amplified. Finally, our improved model adds W++ as a suffix, based on a baseline model shown by the prefix, with the algorithm detailed in Algorithm~\ref{alg3:dwt}.

\begin{algorithm}[t]
	\caption{W++ Sampling.}
	\begin{algorithmic}[1]
		\State \textbf{Input}: Frequency regulation scaling scale $w^f_{t-1}$\,.
		\State $\x_T \sim \mathcal{N}(0, \I)$
		\For{$t = T-1, \dots, 1,0$}
		\State $\z \sim \mathcal{N}(0, \I)$ if $t > 1$, else $\z = 0$
		\State $\x_{t-1} = \frac{1}{\sqrt{\alpha_t}} \left( \x_t - \frac{1 - \alpha_t}{\sqrt{1 - \bar{\alpha}_t}} {\eps_\The(\x_t, t)} \right) + \sigma_t \z$
		\State $\x_{t-1}^{ll}, \x_{t-1}^{lh}, \x_{t-1}^{hl}, \x_{t-1}^{hh} = {\rm \textrm{DWT}}(\x_{t-1})$
		\State $\x_{t-1} = {\rm \textrm{iDWT}}\big(w^f_{t-1}\x_{t-1}^f|f\in\{ll,lh,hl,ll\}\big)$
		\EndFor
		\State \textbf{return} $\x_0$
	\end{algorithmic}
	\label{alg3:dwt}
\end{algorithm}

\section{Experiments}\label{sec:exp}
In this section, we conduct extensive experimental tests and present detailed experimental setups, test results, experimental analyses, and ablation comparisons. Meanwhile, we introduce a large number of improved models targeting exposure bias for comparison. We emphasize that our method can significantly mitigate exposure bias and thereby improve the generation quality of DPMs.

To demonstrate the effectiveness, versatility, and practicality of our methods, we choose a variety of classical diffusion models and samplers, such as DDPM~\cite{ho2020denoising}, IDDPM~\cite{nichol2021improved}, ADM~\cite{dhariwal2021diffusion}, DDIM~\cite{songdenoising}, A-DPM~\cite{baoanalytic}, EA-DPM~\cite{bao2022estimating}, EDM~\cite{Karras2022edm}, PFGM++~\cite{xu2023pfgm++}, and AMED~\cite{zhou2024fast}. Without loss of generality, datasets with different resolutions are selected, such as CIFAR-10~\cite{krizhevsky2009learning}, ImageNet $64\times64$, ImageNet $128\times128$~\cite{chrabaszcz2017downsampled}, and LSUN bedroom $256\times256$~\cite{yu2015lsun}. In general, our experiments are divided into two categories: stochastic generation~\cite{ho2020denoising} and deterministic generation~\cite{song2021scorebased}. Meanwhile, we select different sampling steps for each type of experiment. To more accurately evaluate the generation quality, we select a variety of evaluation metrics, including Fr\'echet Inception Distance (FID)~\cite{heusel2017gans}, Inception Score (IS)~\cite{salimans2016improved}, recall, and precision~\cite{heusel2017gans}. For all metric reports, 50k generated samples are used, with the complete training set as the reference batch. To more intuitively demonstrate the effectiveness of W++, we also present qualitative displays of the generated images. Specifically, we highlight that our research perspective and improved methods aim to address the gap in the study of exposure bias. Thus, we choose some improved models for exposure bias as the comparison models, such as La-DDPM~\cite{zhang2023lookahead}, ADM-IP~\cite{li2023error}, MDSS~\cite{ren2024multi} ADM-ES~\cite{ningelucidating}, and TS-DPM~\cite{li2024alleviating}. To demonstrate the synergy and robustness of our method, we have conducted a large number of ablation experiments, including comparative analysis between single methods and the integrated method, hyperparameter sensitivity tests, and evaluation of computational cost.

\begin{table}[t]
	\centering
	\caption{FID on CIFAR-10 and ImageNet using ADM.}
	\begin{tabularx}{0.45\textwidth}{lcccccc}  
		\toprule
		& \multicolumn{3}{c}{CIFAR-10} & \multicolumn{3}{c}{ImageNet} \\
		\cmidrule(lr){2-4} \cmidrule(lr){5-7} 
		$T'$ & 20 & 30 & 50 & 20 & 30 & 50 \\  
		\midrule
		ADM      & 10.36 & 6.57 & 4.43 & 10.96 & 6.34 & 3.75 \\
		ADM-IP   & 6.89  & 4.25 & 2.92 & -  & - & - \\
		ADM-ES   & 5.15  & 3.37 & 2.49 & 7.52  & 4.63 & 3.07 \\
		\textbf{ADM-W++} & \textbf{4.56} & \textbf{2.65} & \textbf{2.25} &\textbf{7.18} & \textbf{4.30} & \textbf{2.83}  \\
		\bottomrule
	\end{tabularx}
	\label{tab2:adm}
\end{table}

\begin{table}[t]
	\caption{FID on CIFAR-10 using DDPM and IDDPM.}
	\begin{tabular}{lcc lcc}
		\toprule
		Method & 10 & 20 & Method & 30 & 100 \\
		\midrule
		DDPM & 42.04 & 24.60 & IDDPM & 7.81 & 3.72 \\
		TS-DPM & 33.36 & 22.21 & MDSS & 5.25 & 3.49 \\
		\textbf{DDPM-W++} & \textbf{13.54} & \textbf{7.48} & \textbf{IDDPM-W++} & \textbf{3.84} & \textbf{2.77} \\
		\bottomrule
	\end{tabular}
	\label{tab3:ddpm_iddpm}
\end{table}

\subsection{Results on ADM, DDPM, and IDDPM}
To evaluate the versatility of W++, we take ADM~\cite{dhariwal2021diffusion} as the baseline model, which made DPMs surpass GANs by introducing
classifier guidance. Meanwhile, we take ADM-IP~\cite{ning2023input} and ADM-ES~\cite{ningelucidating} as the comparison models, which mitigated exposure bias from the two perspectives of training and sampling, respectively. We conduct unconditional generation on the CIFAR-10 $32\times32$~\cite{krizhevsky2009learning} dataset and class-conditional generation on the Image-Net $64\times64$~\cite{chrabaszcz2017downsampled} dataset with different sampling timesteps.

Table~\ref{tab2:adm} illustrates ADM-W++ achieves markedly superior performance compared to ADM, ADM-IP, and ADM-ES, regardless of datasets or timesteps. It is worth noting that in the 20-step and 30-step sampling tasks on CIFAR-10, ADM-W++ attains the remarkable FID, which is less than One half of ADM’s. Specifically, ``-" in tables indicates that the original paper of the comparison models did not report results for the task.

To prove the superiority of W++, we select the improved models for exposure bias, TS-DPM~\cite{li2024alleviating} and MDSS~\cite{ren2024multi} as the comparison models, with DDPM~\cite{ho2020denoising} and IDDPM~\cite{nichol2021improved} as the baseline models. Table~\ref{tab3:ddpm_iddpm} clearly shows that regardless of the timestep or baseline, W++ achieves much higher generation quality than TS-DPM and MDSS. In particular, in the comparative experiments with TS-DPM, the performance of W++ is far superior to that of TS-DPM, which further highlights the superiority of W++.

To demonstrate the universality of W++, we selected higher-resolution datasets for testing, namely ImageNet $128\times128$ and LSUN bedroom $256\times256$. Meanwhile, we selected ADM and IDDPM as baseline models for these two datasets respectively, to align as closely as possible with the experimental settings of the comparison model described in its original paper. Table~\ref{tab4:image_bed}  makes it clear that, regardless of the dataset or the timestep, W++ achieves better metrics than both the baseline model and the comparison model, which further highlights the generality of W++.

\begin{table}[t]
	\caption{FID on ImageNet $128\times128$ and LSUN Bedroom $256\times256$ using ADM and IDDPM.}
	\begin{tabular}{l c c c c c c}
		\toprule
		& \multicolumn{3}{c}{ADM \& ImageNet} & \multicolumn{3}{c}{IDDPM \& Bedroom} \\
		\cmidrule(lr){2-4} \cmidrule(lr){5-7}
		Method & 20 & 30 & 50 & 20 & 30 & 50 \\
		\midrule
		baseline & 12.48 & 8.04 & 5.15 & 18.63 & 12.99 & 8.50 \\
		baseline-ES & 9.86 & 6.35 & 4.33 & 10.36 & 6.69 & 4.70 \\
		\textbf{baseline-W++} & \textbf{8.81} & \textbf{5.88} & \textbf{4.22} & \textbf{9.37} & \textbf{6.24} & \textbf{4.66} \\
		\bottomrule
	\end{tabular}
	\label{tab4:image_bed}
\end{table}

\begin{table}[t] 
	\centering
	\caption{FID on CIFAR-10 using A-DPM and EA-DPM.}
	\resizebox{0.46\textwidth}{!}{
			\begin{tabular}{lcccccc}
				\toprule
				& \multicolumn{3}{c}{CIFAR10 (LS)} & \multicolumn{3}{c}{CIFAR10 (CS)} \\
				\cmidrule(lr){2-4} \cmidrule(lr){5-7}
				$T'$ & 10 & 25 & 50 & 10 & 25 & 50 \\
				\midrule
				DDPM, $\tilde{\beta}_n$ & 44.45 & 21.83 & 15.21 & 34.76 & 16.18 & 11.11 \\
				DDPM, $\beta_n$ & 233.41 & 125.05 & 66.28 & 205.31 & 84.71 & 37.35 \\
				\midrule
				A-DPM & 34.26 & 11.60 & 7.25 & 22.94 & 8.50 & 5.50 \\
				\textbf{A-DPM-W++} & \textbf{12.38} & \textbf{6.63} & \textbf{4.52} & \textbf{11.61} & \textbf{4.40} & \textbf{3.62} \\
				\midrule
				NPR-DPM & 32.35 & 10.55 & 6.18 & 19.94 & 7.99 & 5.31 \\
				LA-NPR-DPM & 25.59 & 8.84 & 5.28 & - & - & - \\
				\textbf{NPR-DPM-W++} & \textbf{10.86} & \textbf{5.76} & \textbf{4.11} & \textbf{10.18} & \textbf{4.07} & \textbf{3.44} \\
				\midrule
				SN-DPM & 24.06 & 6.91 & 4.63 & 16.33 & 6.05 & 4.17 \\
				LA-SN-DPM & 19.75 & 5.92 & 4.31 & - & - & - \\
				\textbf{SN-DPM-W++} & \textbf{11.73} & \textbf{4.73} & \textbf{3.78} & \textbf{12.53} & \textbf{4.51} & \textbf{3.47} \\
				\bottomrule
			\end{tabular}
		}
		\label{tab5:adpm_eadpm}
	\end{table}
	
	\subsection{Results on A-DPM and EA-DPM}
	To further evaluate the versatility of W++, we choose the improved diffusion models A-DPM (Analytic-DPM)~\cite{baoanalytic} and EA-DPM (Extended-Analytic-DPM)~\cite{bao2022estimating}. The former deduced the analytic form of the optimal reverse variance and the latter used neural networks to estimate the optimal covariance of the conditional Gaussian distribution in the inverse process. 
	Specifically, we tested two advanced models in EA-DPM: NPR-DPM and SN-DPM, where the ``NPR'' and ``SN" symbols represent two different methods for estimating optimal variance under different conditions in EA-DPM. Meanwhile, we conducted various training strategies, employing the linear schedule of $\beta_n$ (LS)~\cite{ho2020denoising}, cosine schedule of $\beta_n$ (CS)~\cite{nichol2021improved} respectively. Finally, we chose LA-DDPM~\cite{zhang2023lookahead} as comparison model because this work made improvements on EA-DPM.
	
	Table~\ref{tab5:adpm_eadpm} shows that W++ outperforms all baseline and comparison models by achieving lower FID scores, regardless of the choices of estimation method, training strategy, or sampling timestep, which not only shows W++ can significantly improve generation quality, but also that W++ is more advanced than other improvements. Similar to the pattern observed with ADM, W++ shows the same degree of performance on any timestep and dataset. These results further confirm the effectiveness and versatility of W++.

	\subsection{Results on DDIM, EDM, FPGM++, \& AMED}
	To further verify the effectiveness and versatility of W++, we select several deterministic sampling models DDIM~\cite{songdenoising}, EDM~\cite{Karras2022edm}, FPGM++~\cite{xu2023pfgm++}, and AMED~\cite{zhou2024fast}. We choose Analytic-DPM~\cite{baoanalytic} as the backbone model for DDIM and the Euler method as the solver for EDM and PFGM++. Unlike the noise prediction network, EDM's neural network restores the original data, PFGM++ is an improved model of the Poisson Flow Generative Model~\cite{xu2022poisson}, and AMED significantly mitigates truncation errors using the median theorem, all of which are crucial for testing the generality of W++. In particular, for DDIM, we use the usual sampling time step as $T'$ as in previous experiments, but for EDM, PFGM ++ and AMED, we choose Neural Function Evaluations (NFE)~\cite{vahdat2021score} as $T'$.
	
	Table~\ref{tab6:ddim_amed} clearly shows that W++ can significantly improve the generation quality of DDIM. For AMED, using the mean value theorem and distillation techniques, W++ can still reduce FID by mitigating exposure bias. Although the EDM illustrates the design space of the diffusion models and PFGM++ is a Poisson Flow Generation Model, the results in Table~\ref{tab7:edm_pfgm} also maintain the same trend, with W++ showing better generation performance than the baseline model and the comparison model. The above experiments show for different deterministic sampling methods, different time steps, and different neural network types, W++ can always significantly improve the generation quality of baseline models, which further proves the effectiveness and versatility of W++.
	
	Particularly, in all the above experiments, $t_{\rm mid}$ is selected as 0.2.
	
	\begin{table}[t]
		\centering
		\caption{\textsc{FID on CIFAR-10 using DDIM and AMED.}}
		\begin{tabularx}{0.46\textwidth}{lXXXXXX}  
			\toprule
			& \multicolumn{3}{c}{DDIM} & \multicolumn{3}{c}{AMED} \\
			\cmidrule(lr){2-4} \cmidrule(lr){5-7}
			$T'$ & 10 & 25 & 50 & 5 & 7 & 9 \\
			Baseline & 26.43 & 9.96 & 6.02 & 7.59 & 4.36 & 3.67 \\
			Baseline-ES & 22.64 & 7.24 & 4.58 & 7.53 & 4.36 & 3.66 \\
			\textbf{Baseline-W++} & \textbf{19.50} & \textbf{6.50} & \textbf{4.46} & \textbf{7.01} & \textbf{4.22} & \textbf{3.54} \\
			\bottomrule
		\end{tabularx}
		\label{tab6:ddim_amed}
	\end{table}
	
	\begin{table}[t]
		\centering
		\caption{\textsc{FID on CIFAR-10 using EDM and PFGM++.}}
		\begin{tabularx}{0.46\textwidth}{lXXXXXX}  
			\toprule
			& \multicolumn{3}{c}{EDM} & \multicolumn{3}{c}{PFGM++} \\
			\cmidrule(lr){2-4} \cmidrule(lr){5-7}
			$T'$ & 13 & 21 & 35 & 13 & 21 & 35 \\
			Baseline & 10.66 & 5.91 & 3.74 & 12.92 & 6.53 & 3.88 \\
			Baseline-ES & 6.59 & 3.74 & 2.59 & 8.79 & 4.54 & 2.91 \\
			\textbf{Baseline-W++} & \textbf{4.68} & \textbf{2.84} & \textbf{2.13} & \textbf{6.62} & \textbf{3.67} & \textbf{2.53} \\
			\bottomrule
		\end{tabularx}
		\label{tab7:edm_pfgm}
	\end{table}

	\begin{table}[t]
		\centering
		\caption{Ablation study of the two regulation schemes.}
		\begin{tabular*}{0.46\textwidth}{@{\extracolsep{\fill}}lcccc@{}}  
			\toprule
			Models & FID$\downarrow$ & IS$\uparrow$ & Precision$\uparrow$ & Recall$\uparrow$ \\  
			\midrule
			ADM & 10.55 & 8.96 & 0.65 & 0.53 \\         
			ADM-low & 7.49 & 9.24 & 0.66 & 0.57\\ 
			ADM-high & 7.58 & 9.67 & 0.66 & 0.57 \\
			\textbf{ADM-W++} & \textbf{4.62} & \textbf{9.74} & \textbf{0.67} & \textbf{0.60} \\ 
			\bottomrule
		\end{tabular*}
		\label{tab7:ablation_study}
	\end{table}
	\begin{table}[t]
		\centering
		\caption{Time for a batch generation on LSUN bedroom $256\times256$ using one GeForce RTX 3090.}
		\begin{tabular}{@{\extracolsep{\fill}}lcccc@{}}  
			\toprule
			Type & IDDPM & IDDPM-W++ & One W++ denoising & One W++ \\  
			\midrule
			Time & 17.35 & 17.42 & 0.873 & 0.003 \\        
			\bottomrule
		\end{tabular}
		\label{tab9:generation_time}
	\end{table}
	\subsection{Qualitative Comparison}
\begin{figure}[t] 
		\centering
		\begin{subfigure}[b]{0.11\textwidth} 
			\includegraphics[width=\linewidth]{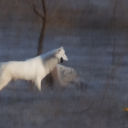}
		\end{subfigure}
		\begin{subfigure}[b]{0.11\textwidth}
			\includegraphics[width=\linewidth]{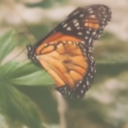}
		\end{subfigure}
		\begin{subfigure}[b]{0.11\textwidth}
			\includegraphics[width=\linewidth]{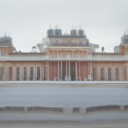}
		\end{subfigure}
		\begin{subfigure}[b]{0.11\textwidth}
			\includegraphics[width=\linewidth]{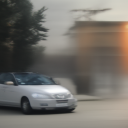}
		\end{subfigure}
		\begin{subfigure}[b]{0.11\textwidth}
			\includegraphics[width=\linewidth]{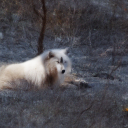}
		\end{subfigure}
		\begin{subfigure}[b]{0.11\textwidth}
			\includegraphics[width=\linewidth]{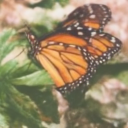}
		\end{subfigure}
		\begin{subfigure}[b]{0.11\textwidth}
			\includegraphics[width=\linewidth]{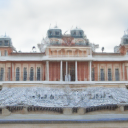}
		\end{subfigure}
		\begin{subfigure}[b]{0.11\textwidth}
			\includegraphics[width=\linewidth]{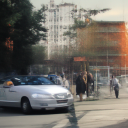}
		\end{subfigure}
		\caption{Qualitative comparison between ADM (first row) and ADM-W++ (second row) using 20 steps.}
		\label{fig7:im_results}
	\end{figure}

	To visually demonstrate the impact of W++ on the generation quality of DPMs, we use the same random seed and sampling steps for both ADM and ADM-W++ to ensure their sampling trajectories are as consistent as possible. As shown in Fig.~\ref{fig7:im_results}, ADM suffers from exposure bias, leading to unnatural artifacts such as over-smoothing, over-exposure, or overly dark regions. In contrast, ADM-W++ effectively mitigates these issues, producing more natural-looking images. In particular, ADM-W++ addresses the frequency energy loss caused by exposure bias by dynamically compensating both low- and high-frequency components, further highlighting the superiority of W++.
	
	\begin{figure}[!t]
		\centering
		\begin{subfigure}[b]{0.495\linewidth}
			\centering
			\includegraphics[width=\linewidth]{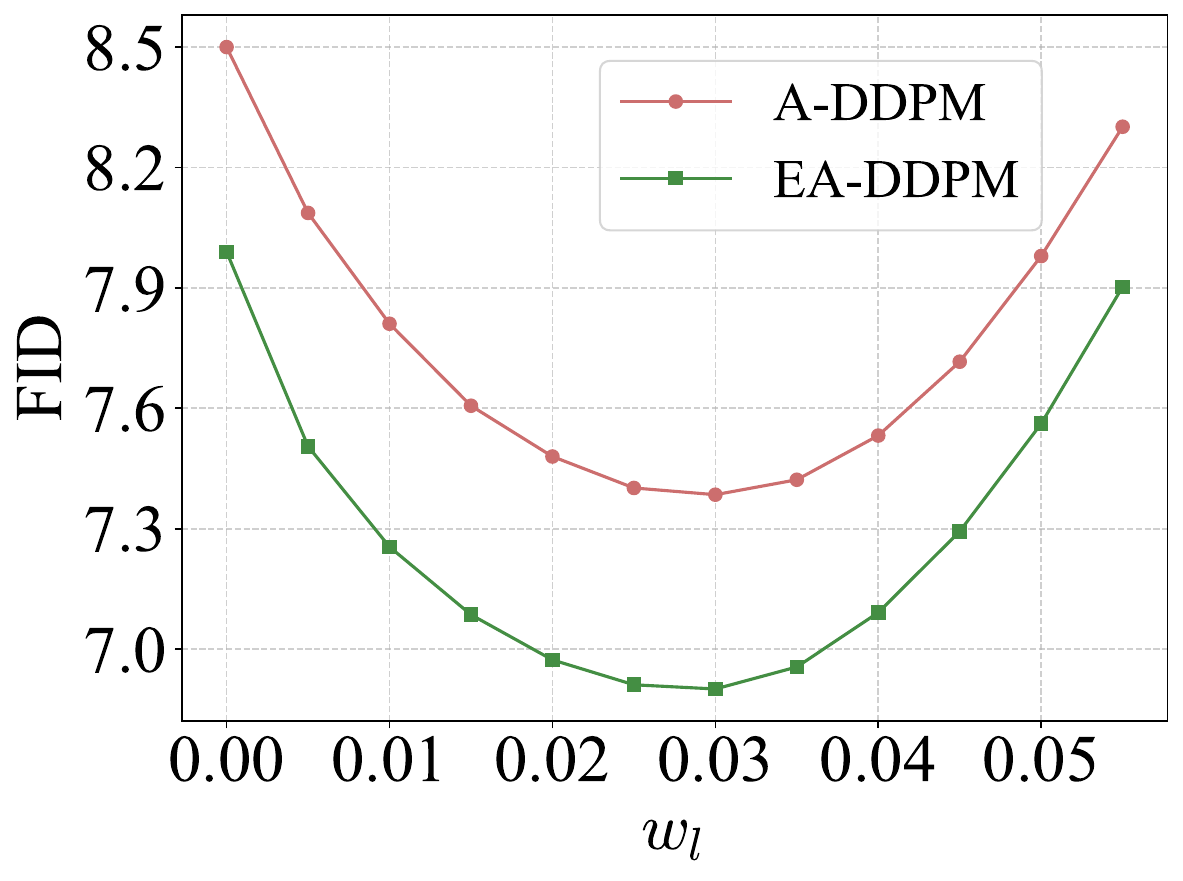}
			\caption{Search experiments of $w_l$\,.}
			\label{fig3a:s}
		\end{subfigure}%
		\begin{subfigure}[b]{0.495\linewidth}
			\centering
			\includegraphics[width=\linewidth]{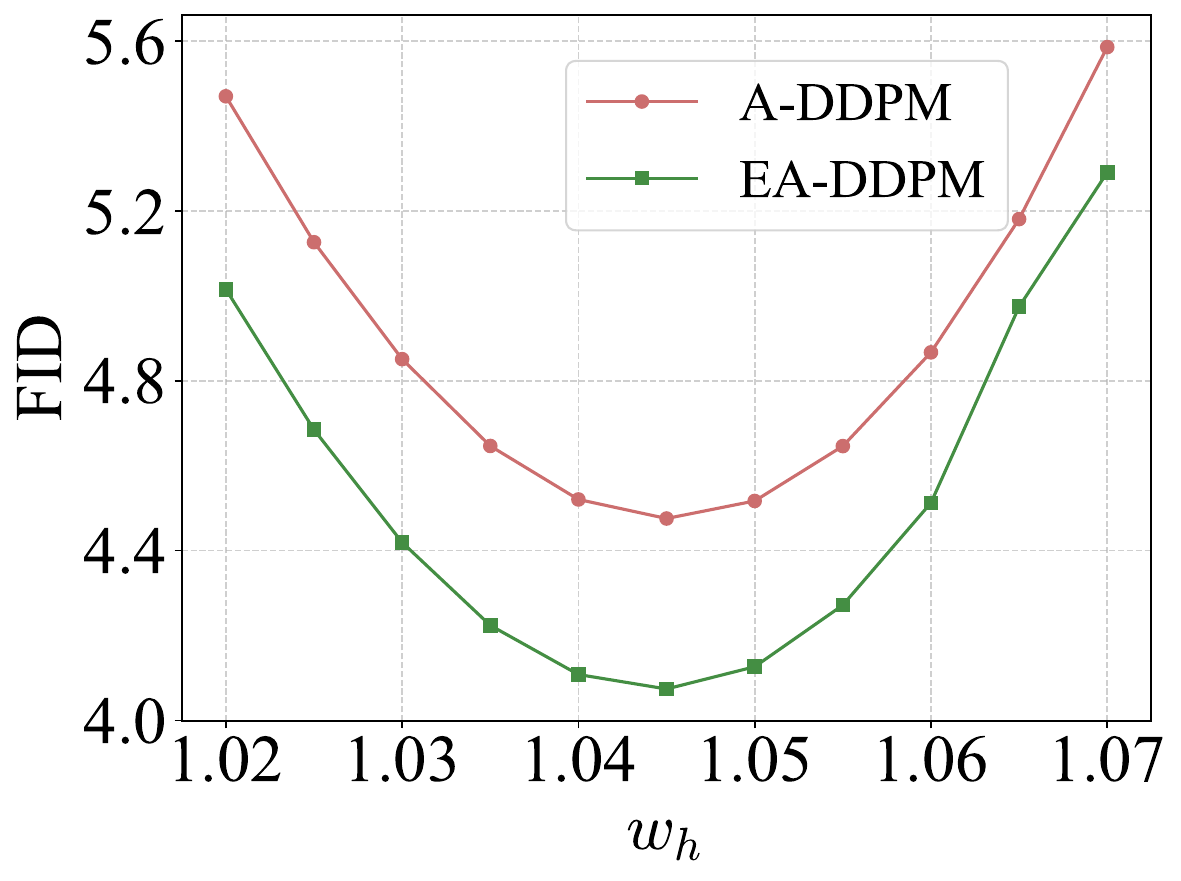}
			\caption{Search experiments of $w_h$\,.}
			\label{fig3b:k}
		\end{subfigure}%
		\caption{Hyperparameters search experiments on CIFAR-10 using A-DPM and EA-DPM with 25 steps.}
		\label{fig8:parameter}
	\end{figure}

	\subsection{Ablation Study}
	\textbf{Synergistic Effect.} 
	To explore the synergy of W++, we separately adjusted the low and high-frequency components, denoted as ADM-low and ADM-high, respectively, and then adjusted both components simultaneously, denoted as ADM-W++. For a comprehensive evaluation of fidelity, diversity, and accuracy, we selected Inception Score (IS), recall, and precision as additional metrics. As shown in Table~\ref{tab7:ablation_study}, adjusting any single frequency component, either low or high, substantially enhances the generation quality. However, the joint adjustment of both low and high-frequency components consistently outperforms individual adjustments, highlighting the synergistic effect of combining both frequency subband.
	
	\textbf{Generation Delay.}
	To examine the impact of introducing W++ on sampling time, we fairly compared the time for generating a batch of samples between IDDPM and IDDPM-W++ under the same hardware, random number seed, and dataset. Meanwhile, we repeated the experiment $100$ times to calculate the average time. Table~\ref{tab9:generation_time} clearly shows IDDPM-W++ takes only $0.007$ seconds longer than IDDPM to generate one batch, with an approximately $0.4\%$ increase in sampling time, which is nearly negligible. Meanwhile, the W++ operation takes only $0.003$ seconds, accounting for approximately $0.3\%$ of the single-step denoising time, which further confirms that W++ does not cause generation delays.

	\textbf{Hyperparameter Insensitivity.} 
	To demonstrate the insensitivity of W++ to hyperparameters, we present the detailed process of parameter search. Initially, we adjusted only the low-frequency component of baselines to find the optimal parameter \(w_l^*\). Then, based on the optimal \(w_l^*\), we fine-tuned the high-frequency components to determine the optimal \(w_h^*\). Fig.~\ref{fig8:parameter} clearly shows that W++ achieves consistent performance improvements over a wide range of \(w_l\) and \(w_h\), which indicates the method’s insensitivity to hyperparameters. Notably, Fig.~\ref{fig8:parameter} also shows the FID curves always follow the pattern of first decreasing and then increasing, and the inflection point corresponds to the optimal parameter. This explicit characteristic of the minimum point enables us to quickly lock in the optimal parameter, further verifying the practicality of W++.

	\section{Conclusion}\label{sec:con}
	To the best of our knowledge, we are the first to analyze and mitigate the exposure bias of diffusion probabilistic models in the wavelet domain. Our analysis reveals that the subband energy of predicted samples in the reverse process follows distinct reduction patterns. To address this, we propose a simple yet effective frequency regulation mechanism that dynamically enhances both low- and high-frequency subbands during sampling. Furthermore, we observe that the subband energy of reconstructed samples is consistently lower than that of the original data, providing strong evidence for our theoretical analysis and enabling us to derive the analytical form of exposure bias. Importantly, our method is training-free, plug-and-play, and improves the generation quality of diverse diffusion probabilistic models and frameworks with negligible computational overhead. Extensive experiments across datasets of varying resolutions, samplers, and time steps consistently demonstrate that our approach achieves superior performance over existing exposure bias mitigation methods, underscoring its effectiveness and generality.
\begin{acks}
This work was supported by the National Natural Science Foundation of China under the Grant No. 62176108 and the Supercomputing Center of Lanzhou University.
\end{acks}
	
	\bibliographystyle{ACM-Reference-Format}
	\balance
	\bibliography{bias}


\begin{thebibliography}{51}


\ifx \showCODEN    \undefined \def \showCODEN     #1{\unskip}     \fi
\ifx \showISBNx    \undefined \def \showISBNx     #1{\unskip}     \fi
\ifx \showISBNxiii \undefined \def \showISBNxiii  #1{\unskip}     \fi
\ifx \showISSN     \undefined \def \showISSN      #1{\unskip}     \fi
\ifx \showLCCN     \undefined \def \showLCCN      #1{\unskip}     \fi
\ifx \shownote     \undefined \def \shownote      #1{#1}          \fi
\ifx \showarticletitle \undefined \def \showarticletitle #1{#1}   \fi
\ifx \showURL      \undefined \def \showURL       {\relax}        \fi
\providecommand\bibfield[2]{#2}
\providecommand\bibinfo[2]{#2}
\providecommand\natexlab[1]{#1}
\providecommand\showeprint[2][]{arXiv:#2}

\bibitem[Bao et~al\mbox{.}(2022a)]%
        {bao2022estimating}
\bibfield{author}{\bibinfo{person}{Fan Bao}, \bibinfo{person}{Chongxuan Li},
  \bibinfo{person}{Jiacheng Sun}, \bibinfo{person}{Jun Zhu}, {and}
  \bibinfo{person}{Bo Zhang}.} \bibinfo{year}{2022}\natexlab{a}.
\newblock \showarticletitle{Estimating the Optimal Covariance with Imperfect
  Mean in Diffusion Probabilistic Models}. In \bibinfo{booktitle}{\emph{ICML}}.
\newblock


\bibitem[Bao et~al\mbox{.}(2022b)]%
        {baoanalytic}
\bibfield{author}{\bibinfo{person}{Fan Bao}, \bibinfo{person}{Chongxuan Li},
  \bibinfo{person}{Jun Zhu}, {and} \bibinfo{person}{Bo Zhang}.}
  \bibinfo{year}{2022}\natexlab{b}.
\newblock \showarticletitle{{Analytic-DPM}: an Analytic Estimate of the Optimal
  Reverse Variance in Diffusion Probabilistic Models}. In
  \bibinfo{booktitle}{\emph{International Conference on Learning
  Representations}}.
\newblock


\bibitem[Chrabaszcz et~al\mbox{.}(2017)]%
        {chrabaszcz2017downsampled}
\bibfield{author}{\bibinfo{person}{Patryk Chrabaszcz}, \bibinfo{person}{Ilya
  Loshchilov}, {and} \bibinfo{person}{Frank Hutter}.}
  \bibinfo{year}{2017}\natexlab{}.
\newblock \showarticletitle{A downsampled variant of {ImageNet} as an
  alternative to the {CIFAR} datasets}.
\newblock \bibinfo{journal}{\emph{arXiv preprint arXiv:1707.08819}}
  (\bibinfo{year}{2017}).
\newblock


\bibitem[Dhariwal and Nichol(2021)]%
        {dhariwal2021diffusion}
\bibfield{author}{\bibinfo{person}{Prafulla Dhariwal} {and}
  \bibinfo{person}{Alexander Nichol}.} \bibinfo{year}{2021}\natexlab{}.
\newblock \showarticletitle{Diffusion models beat {GANs} on image synthesis}.
  In \bibinfo{booktitle}{\emph{NeurIPS}}.
\newblock


\bibitem[Dockhorn et~al\mbox{.}(2022)]%
        {dockhorn2022genie}
\bibfield{author}{\bibinfo{person}{Tim Dockhorn}, \bibinfo{person}{Arash
  Vahdat}, {and} \bibinfo{person}{Karsten Kreis}.}
  \bibinfo{year}{2022}\natexlab{}.
\newblock \showarticletitle{Genie: Higher-order denoising diffusion solvers}.
  In \bibinfo{booktitle}{\emph{NeurIPS}}.
\newblock


\bibitem[Gal et~al\mbox{.}(2021)]%
        {gal2021swagan}
\bibfield{author}{\bibinfo{person}{Rinon Gal}, \bibinfo{person}{Dana~Cohen
  Hochberg}, \bibinfo{person}{Amit Bermano}, {and} \bibinfo{person}{Daniel
  Cohen-Or}.} \bibinfo{year}{2021}\natexlab{}.
\newblock \showarticletitle{Swagan: A style-based wavelet-driven generative
  model}.
\newblock \bibinfo{journal}{\emph{ACM Transactions on Graphics}}
  (\bibinfo{year}{2021}).
\newblock


\bibitem[Goodfellow et~al\mbox{.}(2014)]%
        {goodfellow2014generative}
\bibfield{author}{\bibinfo{person}{Ian Goodfellow}, \bibinfo{person}{Jean
  Pouget-Abadie}, \bibinfo{person}{Mehdi Mirza}, \bibinfo{person}{Bing Xu},
  \bibinfo{person}{David Warde-Farley}, \bibinfo{person}{Sherjil Ozair},
  \bibinfo{person}{Aaron Courville}, {and} \bibinfo{person}{Yoshua Bengio}.}
  \bibinfo{year}{2014}\natexlab{}.
\newblock \showarticletitle{Generative adversarial nets}. In
  \bibinfo{booktitle}{\emph{NeurIPS}}.
\newblock


\bibitem[Graps(1995)]%
        {graps1995introduction}
\bibfield{author}{\bibinfo{person}{Amara Graps}.}
  \bibinfo{year}{1995}\natexlab{}.
\newblock \showarticletitle{An introduction to wavelets}.
\newblock \bibinfo{journal}{\emph{IEEE Computational Science and Engineering}}
  (\bibinfo{year}{1995}).
\newblock


\bibitem[Guth et~al\mbox{.}(2022)]%
        {guth2022wavelet}
\bibfield{author}{\bibinfo{person}{Florentin Guth}, \bibinfo{person}{Simon
  Coste}, \bibinfo{person}{Valentin De~Bortoli}, {and}
  \bibinfo{person}{Stephane Mallat}.} \bibinfo{year}{2022}\natexlab{}.
\newblock \showarticletitle{Wavelet score-based generative modeling}. In
  \bibinfo{booktitle}{\emph{NeurIPS}}.
\newblock


\bibitem[Heusel et~al\mbox{.}(2017)]%
        {heusel2017gans}
\bibfield{author}{\bibinfo{person}{Martin Heusel}, \bibinfo{person}{Hubert
  Ramsauer}, \bibinfo{person}{Thomas Unterthiner}, \bibinfo{person}{Bernhard
  Nessler}, {and} \bibinfo{person}{Sepp Hochreiter}.}
  \bibinfo{year}{2017}\natexlab{}.
\newblock \showarticletitle{{GANs} trained by a two time-scale update rule
  converge to a local nash equilibrium}. In
  \bibinfo{booktitle}{\emph{NeurIPS}}.
\newblock


\bibitem[Ho et~al\mbox{.}(2020)]%
        {ho2020denoising}
\bibfield{author}{\bibinfo{person}{Jonathan Ho}, \bibinfo{person}{Ajay Jain},
  {and} \bibinfo{person}{Pieter Abbeel}.} \bibinfo{year}{2020}\natexlab{}.
\newblock \showarticletitle{Denoising Diffusion Probabilistic Models}. In
  \bibinfo{booktitle}{\emph{NeurIPS}}.
\newblock


\bibitem[Karras et~al\mbox{.}(2022)]%
        {Karras2022edm}
\bibfield{author}{\bibinfo{person}{Tero Karras}, \bibinfo{person}{Miika
  Aittala}, \bibinfo{person}{Timo Aila}, {and} \bibinfo{person}{Samuli Laine}.}
  \bibinfo{year}{2022}\natexlab{}.
\newblock \showarticletitle{Elucidating the Design Space of Diffusion-Based
  Generative Models}. In \bibinfo{booktitle}{\emph{NeurIPS}}.
\newblock


\bibitem[Kim et~al\mbox{.}(2023)]%
        {kim2023refining}
\bibfield{author}{\bibinfo{person}{Dongjun Kim}, \bibinfo{person}{Yeongmin
  Kim}, \bibinfo{person}{Se~Jung Kwon}, \bibinfo{person}{Wanmo Kang}, {and}
  \bibinfo{person}{Il-Chul Moon}.} \bibinfo{year}{2023}\natexlab{}.
\newblock \showarticletitle{Refining Generative Process with Discriminator
  Guidance in Score-based Diffusion Models}. In
  \bibinfo{booktitle}{\emph{ICML}}.
\newblock


\bibitem[Krizhevsky et~al\mbox{.}(2009)]%
        {krizhevsky2009learning}
\bibfield{author}{\bibinfo{person}{Alex Krizhevsky}, \bibinfo{person}{Geoffrey
  Hinton}, {et~al\mbox{.}}} \bibinfo{year}{2009}\natexlab{}.
\newblock \showarticletitle{Learning multiple layers of features from tiny
  images}.
\newblock  (\bibinfo{year}{2009}).
\newblock


\bibitem[Li et~al\mbox{.}(2022)]%
        {li2022wavelet}
\bibfield{author}{\bibinfo{person}{Jin Li}, \bibinfo{person}{Wanyun Li},
  \bibinfo{person}{Zichen Xu}, \bibinfo{person}{Yuhao Wang}, {and}
  \bibinfo{person}{Qiegen Liu}.} \bibinfo{year}{2022}\natexlab{}.
\newblock \showarticletitle{Wavelet transform-assisted adaptive generative
  modeling for colorization}.
\newblock \bibinfo{journal}{\emph{IEEE Transactions on Multimedia}}
  (\bibinfo{year}{2022}).
\newblock


\bibitem[Li et~al\mbox{.}(2024)]%
        {li2024alleviating}
\bibfield{author}{\bibinfo{person}{Mingxiao Li}, \bibinfo{person}{Tingyu Qu},
  \bibinfo{person}{Ruicong Yao}, \bibinfo{person}{Wei Sun}, {and}
  \bibinfo{person}{Marie-Francine Moens}.} \bibinfo{year}{2024}\natexlab{}.
\newblock \showarticletitle{Alleviating Exposure Bias in Diffusion Models
  through Sampling with Shifted Time Steps}. In
  \bibinfo{booktitle}{\emph{ICLR}}.
\newblock


\bibitem[Li and van~der Schaar(2024)]%
        {li2023error}
\bibfield{author}{\bibinfo{person}{Yangming Li} {and} \bibinfo{person}{Mihaela
  van~der Schaar}.} \bibinfo{year}{2024}\natexlab{}.
\newblock \showarticletitle{On Error Propagation of Diffusion Models}. In
  \bibinfo{booktitle}{\emph{ICLR}}.
\newblock


\bibitem[Liu et~al\mbox{.}(2023)]%
        {liu2023instaflow}
\bibfield{author}{\bibinfo{person}{Xingchao Liu}, \bibinfo{person}{Xiwen
  Zhang}, \bibinfo{person}{Jianzhu Ma}, \bibinfo{person}{Jian Peng},
  {et~al\mbox{.}}} \bibinfo{year}{2023}\natexlab{}.
\newblock \showarticletitle{Instaflow: One step is enough for high-quality
  diffusion-based text-to-image generation}. In
  \bibinfo{booktitle}{\emph{ICLR}}.
\newblock


\bibitem[Lu and Song(2025)]%
        {lu2025simplifying}
\bibfield{author}{\bibinfo{person}{Cheng Lu} {and} \bibinfo{person}{Yang
  Song}.} \bibinfo{year}{2025}\natexlab{}.
\newblock \showarticletitle{Simplifying, Stabilizing and Scaling
  Continuous-time Consistency Models}. In \bibinfo{booktitle}{\emph{ICLR}}.
\newblock


\bibitem[Lu et~al\mbox{.}(2022)]%
        {lu2022dpmsolver}
\bibfield{author}{\bibinfo{person}{Cheng Lu}, \bibinfo{person}{Yuhao Zhou},
  \bibinfo{person}{Fan Bao}, \bibinfo{person}{Jianfei Chen},
  \bibinfo{person}{Chongxuan Li}, {and} \bibinfo{person}{Jun Zhu}.}
  \bibinfo{year}{2022}\natexlab{}.
\newblock \showarticletitle{{DPM}-Solver: A Fast {ODE} Solver for Diffusion
  Probabilistic Model Sampling in Around 10 Steps}. In
  \bibinfo{booktitle}{\emph{NeurIPS}}.
\newblock


\bibitem[Luhman and Luhman(2021)]%
        {luhman2021knowledge}
\bibfield{author}{\bibinfo{person}{Eric Luhman} {and} \bibinfo{person}{Troy
  Luhman}.} \bibinfo{year}{2021}\natexlab{}.
\newblock \showarticletitle{Knowledge distillation in iterative generative
  models for improved sampling speed}.
\newblock \bibinfo{journal}{\emph{arXiv preprint arXiv:2101.02388}}
  (\bibinfo{year}{2021}).
\newblock


\bibitem[Mallat(1989)]%
        {mallat1989theory}
\bibfield{author}{\bibinfo{person}{Stephane~G Mallat}.}
  \bibinfo{year}{1989}\natexlab{}.
\newblock \showarticletitle{A theory for multiresolution signal decomposition:
  the wavelet representation}.
\newblock \bibinfo{journal}{\emph{IEEE Transactions on Pattern Analysis and
  Machine Intelligence}} (\bibinfo{year}{1989}).
\newblock


\bibitem[Meng et~al\mbox{.}(2023)]%
        {meng2023distillation}
\bibfield{author}{\bibinfo{person}{Chenlin Meng}, \bibinfo{person}{Robin
  Rombach}, \bibinfo{person}{Ruiqi Gao}, \bibinfo{person}{Diederik Kingma},
  \bibinfo{person}{Stefano Ermon}, \bibinfo{person}{Jonathan Ho}, {and}
  \bibinfo{person}{Tim Salimans}.} \bibinfo{year}{2023}\natexlab{}.
\newblock \showarticletitle{On distillation of guided diffusion models}. In
  \bibinfo{booktitle}{\emph{CVPR}}.
\newblock


\bibitem[Nichol and Dhariwal(2021)]%
        {nichol2021improved}
\bibfield{author}{\bibinfo{person}{Alexander~Quinn Nichol} {and}
  \bibinfo{person}{Prafulla Dhariwal}.} \bibinfo{year}{2021}\natexlab{}.
\newblock \showarticletitle{Improved denoising diffusion probabilistic models}.
  In \bibinfo{booktitle}{\emph{ICLR}}.
\newblock


\bibitem[Ning et~al\mbox{.}(2024)]%
        {ningelucidating}
\bibfield{author}{\bibinfo{person}{Mang Ning}, \bibinfo{person}{Mingxiao Li},
  \bibinfo{person}{Jianlin Su}, \bibinfo{person}{Albert~Ali Salah}, {and}
  \bibinfo{person}{Itir~Onal Ertugrul}.} \bibinfo{year}{2024}\natexlab{}.
\newblock \showarticletitle{Elucidating the Exposure Bias in Diffusion Models}.
  In \bibinfo{booktitle}{\emph{ICLR}}.
\newblock


\bibitem[Ning et~al\mbox{.}(2023)]%
        {ning2023input}
\bibfield{author}{\bibinfo{person}{Mang Ning}, \bibinfo{person}{Enver
  Sangineto}, \bibinfo{person}{Angelo Porrello}, \bibinfo{person}{Simone
  Calderara}, {and} \bibinfo{person}{Rita Cucchiara}.}
  \bibinfo{year}{2023}\natexlab{}.
\newblock \showarticletitle{Input Perturbation Reduces Exposure Bias in
  Diffusion Models}. In \bibinfo{booktitle}{\emph{ICML}}.
\newblock


\bibitem[Phung et~al\mbox{.}(2023)]%
        {phung2023wavelet}
\bibfield{author}{\bibinfo{person}{Hao Phung}, \bibinfo{person}{Quan Dao},
  {and} \bibinfo{person}{Anh Tran}.} \bibinfo{year}{2023}\natexlab{}.
\newblock \showarticletitle{Wavelet diffusion models are fast and scalable
  image generators}. In \bibinfo{booktitle}{\emph{CVPR}}.
\newblock


\bibitem[Ren et~al\mbox{.}(2024)]%
        {ren2024multi}
\bibfield{author}{\bibinfo{person}{Zhiyao Ren}, \bibinfo{person}{Yibing Zhan},
  \bibinfo{person}{Liang Ding}, \bibinfo{person}{Gaoang Wang},
  \bibinfo{person}{Chaoyue Wang}, \bibinfo{person}{Zhongyi Fan}, {and}
  \bibinfo{person}{Dacheng Tao}.} \bibinfo{year}{2024}\natexlab{}.
\newblock \showarticletitle{Multi-Step Denoising Scheduled Sampling: Towards
  Alleviating Exposure Bias for Diffusion Models}. In
  \bibinfo{booktitle}{\emph{AAAI}}. \bibinfo{pages}{4667--4675}.
\newblock


\bibitem[Rombach et~al\mbox{.}(2022)]%
        {rombach2022high}
\bibfield{author}{\bibinfo{person}{Robin Rombach}, \bibinfo{person}{Andreas
  Blattmann}, \bibinfo{person}{Dominik Lorenz}, \bibinfo{person}{Patrick
  Esser}, {and} \bibinfo{person}{Bj{\"o}rn Ommer}.}
  \bibinfo{year}{2022}\natexlab{}.
\newblock \showarticletitle{High-resolution image synthesis with latent
  diffusion models}. In \bibinfo{booktitle}{\emph{CVPR}}.
\newblock


\bibitem[Salimans et~al\mbox{.}(2016)]%
        {salimans2016improved}
\bibfield{author}{\bibinfo{person}{Tim Salimans}, \bibinfo{person}{Ian
  Goodfellow}, \bibinfo{person}{Wojciech Zaremba}, \bibinfo{person}{Vicki
  Cheung}, \bibinfo{person}{Alec Radford}, {and} \bibinfo{person}{Xi Chen}.}
  \bibinfo{year}{2016}\natexlab{}.
\newblock \showarticletitle{Improved techniques for training {GANs}}. In
  \bibinfo{booktitle}{\emph{NeurIPS}}.
\newblock


\bibitem[Salimans and Ho(2022)]%
        {salimans2022progressive}
\bibfield{author}{\bibinfo{person}{Tim Salimans} {and}
  \bibinfo{person}{Jonathan Ho}.} \bibinfo{year}{2022}\natexlab{}.
\newblock \showarticletitle{Progressive Distillation for Fast Sampling of
  Diffusion Models}. In \bibinfo{booktitle}{\emph{ICLR}}.
\newblock


\bibitem[Sohl-Dickstein et~al\mbox{.}(2015)]%
        {sohl2015deep}
\bibfield{author}{\bibinfo{person}{Jascha Sohl-Dickstein},
  \bibinfo{person}{Eric Weiss}, \bibinfo{person}{Niru Maheswaranathan}, {and}
  \bibinfo{person}{Surya Ganguli}.} \bibinfo{year}{2015}\natexlab{}.
\newblock \showarticletitle{Deep unsupervised learning using nonequilibrium
  thermodynamics}. In \bibinfo{booktitle}{\emph{ICML}}.
  \bibinfo{pages}{2256--2265}.
\newblock


\bibitem[Song et~al\mbox{.}(2021a)]%
        {songdenoising}
\bibfield{author}{\bibinfo{person}{Jiaming Song}, \bibinfo{person}{Chenlin
  Meng}, {and} \bibinfo{person}{Stefano Ermon}.}
  \bibinfo{year}{2021}\natexlab{a}.
\newblock \showarticletitle{Denoising Diffusion Implicit Models}. In
  \bibinfo{booktitle}{\emph{ICLR}}.
\newblock


\bibitem[Song and Dhariwal(2024)]%
        {song2024improved}
\bibfield{author}{\bibinfo{person}{Yang Song} {and} \bibinfo{person}{Prafulla
  Dhariwal}.} \bibinfo{year}{2024}\natexlab{}.
\newblock \showarticletitle{Improved Techniques for Training Consistency
  Models}. In \bibinfo{booktitle}{\emph{ICLR}}.
\newblock


\bibitem[Song et~al\mbox{.}(2023)]%
        {song2023consistency}
\bibfield{author}{\bibinfo{person}{Yang Song}, \bibinfo{person}{Prafulla
  Dhariwal}, \bibinfo{person}{Mark Chen}, {and} \bibinfo{person}{Ilya
  Sutskever}.} \bibinfo{year}{2023}\natexlab{}.
\newblock \showarticletitle{Consistency models}. In
  \bibinfo{booktitle}{\emph{ICML}}.
\newblock


\bibitem[Song et~al\mbox{.}(2021b)]%
        {song2021scorebased}
\bibfield{author}{\bibinfo{person}{Yang Song}, \bibinfo{person}{Jascha
  Sohl-Dickstein}, \bibinfo{person}{Diederik~P Kingma},
  \bibinfo{person}{Abhishek Kumar}, \bibinfo{person}{Stefano Ermon}, {and}
  \bibinfo{person}{Ben Poole}.} \bibinfo{year}{2021}\natexlab{b}.
\newblock \showarticletitle{Score-Based Generative Modeling through Stochastic
  Differential Equations}. In \bibinfo{booktitle}{\emph{ICLR}}.
\newblock


\bibitem[Vahdat et~al\mbox{.}(2021)]%
        {vahdat2021score}
\bibfield{author}{\bibinfo{person}{Arash Vahdat}, \bibinfo{person}{Karsten
  Kreis}, {and} \bibinfo{person}{Jan Kautz}.} \bibinfo{year}{2021}\natexlab{}.
\newblock \showarticletitle{Score-based generative modeling in latent space}.
\newblock \bibinfo{journal}{\emph{NeurIPS}}.
\newblock


\bibitem[Wang et~al\mbox{.}(2022)]%
        {wang2022fregan}
\bibfield{author}{\bibinfo{person}{Zhe Wang}, \bibinfo{person}{Ziqiu Chi},
  \bibinfo{person}{Yanbing Zhang}, {et~al\mbox{.}}}
  \bibinfo{year}{2022}\natexlab{}.
\newblock \showarticletitle{FreGAN: Exploiting frequency components for
  training GANs under limited data}.
\newblock \bibinfo{journal}{\emph{NeurIPS}} (\bibinfo{year}{2022}).
\newblock


\bibitem[Xu et~al\mbox{.}(2022)]%
        {xu2022poisson}
\bibfield{author}{\bibinfo{person}{Yilun Xu}, \bibinfo{person}{Ziming Liu},
  \bibinfo{person}{Max Tegmark}, {and} \bibinfo{person}{Tommi Jaakkola}.}
  \bibinfo{year}{2022}\natexlab{}.
\newblock \showarticletitle{Poisson flow generative models}. In
  \bibinfo{booktitle}{\emph{NeurIPS}}.
\newblock


\bibitem[Xu et~al\mbox{.}(2023)]%
        {xu2023pfgm++}
\bibfield{author}{\bibinfo{person}{Yilun Xu}, \bibinfo{person}{Ziming Liu},
  \bibinfo{person}{Yonglong Tian}, \bibinfo{person}{Shangyuan Tong},
  \bibinfo{person}{Max Tegmark}, {and} \bibinfo{person}{Tommi Jaakkola}.}
  \bibinfo{year}{2023}\natexlab{}.
\newblock \showarticletitle{{PFGM++}: Unlocking the potential of
  physics-inspired generative models}. In \bibinfo{booktitle}{\emph{ICML}}.
\newblock


\bibitem[Yang et~al\mbox{.}(2022)]%
        {yang2022wavegan}
\bibfield{author}{\bibinfo{person}{Mengping Yang}, \bibinfo{person}{Zhe Wang},
  \bibinfo{person}{Ziqiu Chi}, {and} \bibinfo{person}{Wenyi Feng}.}
  \bibinfo{year}{2022}\natexlab{}.
\newblock \showarticletitle{Wavegan: Frequency-aware gan for high-fidelity
  few-shot image generation}. In \bibinfo{booktitle}{\emph{ECCV}}.
\newblock


\bibitem[YAO et~al\mbox{.}(2025)]%
        {yao2025manifold}
\bibfield{author}{\bibinfo{person}{Yuzhe YAO}, \bibinfo{person}{Jun Chen},
  \bibinfo{person}{Zeyi Huang}, \bibinfo{person}{Haonan Lin},
  \bibinfo{person}{Mengmeng Wang}, \bibinfo{person}{Guang Dai}, {and}
  \bibinfo{person}{Jingdong Wang}.} \bibinfo{year}{2025}\natexlab{}.
\newblock \showarticletitle{Manifold Constraint Reduces Exposure Bias in
  Accelerated Diffusion Sampling}. In \bibinfo{booktitle}{\emph{ICLR}}.
\newblock


\bibitem[Yu et~al\mbox{.}(2015)]%
        {yu2015lsun}
\bibfield{author}{\bibinfo{person}{Fisher Yu}, \bibinfo{person}{Ari Seff},
  \bibinfo{person}{Yinda Zhang}, \bibinfo{person}{Shuran Song},
  \bibinfo{person}{Thomas Funkhouser}, {and} \bibinfo{person}{Jianxiong Xiao}.}
  \bibinfo{year}{2015}\natexlab{}.
\newblock \showarticletitle{Lsun: Construction of a large-scale image dataset
  using deep learning with humans in the loop}.
\newblock \bibinfo{journal}{\emph{arXiv preprint arXiv:1506.03365}}
  (\bibinfo{year}{2015}).
\newblock


\bibitem[Yu and Zhan(2025)]%
        {yubias}
\bibfield{author}{\bibinfo{person}{Meng Yu} {and} \bibinfo{person}{Kun Zhan}.}
  \bibinfo{year}{2025}\natexlab{}.
\newblock \showarticletitle{Bias Mitigation in Graph Diffusion Models}. In
  \bibinfo{booktitle}{\emph{ICLR}}.
\newblock


\bibitem[Zhang et~al\mbox{.}(2022)]%
        {zhang2022styleswin}
\bibfield{author}{\bibinfo{person}{Bowen Zhang}, \bibinfo{person}{Shuyang Gu},
  \bibinfo{person}{Bo Zhang}, \bibinfo{person}{Jianmin Bao},
  \bibinfo{person}{Dong Chen}, \bibinfo{person}{Fang Wen},
  \bibinfo{person}{Yong Wang}, {and} \bibinfo{person}{Baining Guo}.}
  \bibinfo{year}{2022}\natexlab{}.
\newblock \showarticletitle{Styleswin: Transformer-based gan for
  high-resolution image generation}. In \bibinfo{booktitle}{\emph{CVPR}}.
\newblock


\bibitem[Zhang et~al\mbox{.}(2023)]%
        {zhang2023lookahead}
\bibfield{author}{\bibinfo{person}{Guoqiang Zhang}, \bibinfo{person}{Kenta
  Niwa}, {and} \bibinfo{person}{W~Bastiaan Kleijn}.}
  \bibinfo{year}{2023}\natexlab{}.
\newblock \showarticletitle{Lookahead diffusion probabilistic models for
  refining mean estimation}. In \bibinfo{booktitle}{\emph{CVPR}}.
\newblock


\bibitem[Zhang et~al\mbox{.}(2025)]%
        {zhang2025antiexposure}
\bibfield{author}{\bibinfo{person}{Junyu Zhang}, \bibinfo{person}{Daochang
  Liu}, \bibinfo{person}{Eunbyung Park}, \bibinfo{person}{Shichao Zhang}, {and}
  \bibinfo{person}{Chang Xu}.} \bibinfo{year}{2025}\natexlab{}.
\newblock \showarticletitle{Anti-Exposure Bias in Diffusion Models}. In
  \bibinfo{booktitle}{\emph{ICLR}}.
\newblock


\bibitem[Zhang and Chen(2023)]%
        {zhang2023fast}
\bibfield{author}{\bibinfo{person}{Qinsheng Zhang} {and}
  \bibinfo{person}{Yongxin Chen}.} \bibinfo{year}{2023}\natexlab{}.
\newblock \showarticletitle{Fast Sampling of Diffusion Models with Exponential
  Integrator}. In \bibinfo{booktitle}{\emph{ICLR}}.
\newblock


\bibitem[Zhao et~al\mbox{.}(2024)]%
        {zhao2024unipc}
\bibfield{author}{\bibinfo{person}{Wenliang Zhao}, \bibinfo{person}{Lujia Bai},
  \bibinfo{person}{Yongming Rao}, \bibinfo{person}{Jie Zhou}, {and}
  \bibinfo{person}{Jiwen Lu}.} \bibinfo{year}{2024}\natexlab{}.
\newblock \showarticletitle{Unipc: A unified predictor-corrector framework for
  fast sampling of diffusion models}. In \bibinfo{booktitle}{\emph{NeurIPS}}.
\newblock


\bibitem[Zheng et~al\mbox{.}(2023)]%
        {zheng2023fast}
\bibfield{author}{\bibinfo{person}{Hongkai Zheng}, \bibinfo{person}{Weili Nie},
  \bibinfo{person}{Arash Vahdat}, \bibinfo{person}{Kamyar Azizzadenesheli},
  {and} \bibinfo{person}{Anima Anandkumar}.} \bibinfo{year}{2023}\natexlab{}.
\newblock \showarticletitle{Fast sampling of diffusion models via operator
  learning}. In \bibinfo{booktitle}{\emph{ICML}}.
\newblock


\bibitem[Zhou et~al\mbox{.}(2024)]%
        {zhou2024fast}
\bibfield{author}{\bibinfo{person}{Zhenyu Zhou}, \bibinfo{person}{Defang Chen},
  \bibinfo{person}{Can Wang}, {and} \bibinfo{person}{Chun Chen}.}
  \bibinfo{year}{2024}\natexlab{}.
\newblock \showarticletitle{Fast ode-based sampling for diffusion models in
  around 5 steps}. In \bibinfo{booktitle}{\emph{CVPR}}.
\newblock


\end{thebibliography}
\clearpage
\section*{Appendix A}
In this section, we give a detailed proof of the Eq.~\eqref{eq13.bias_reverse}. For ease of understanding, we will rewrite the necessary formulas here and we state that all the $\eps$ proposed in this section follow the standard Gaussian distribution. We know that the perturbed sample in the forward process is
\begin{equation}
	\x_t=\sqrt{\bar{\alpha}_t}\x_0+\sqrt{1-\bar{\alpha}_t}\eps.
	\label{eq20:forward_one}
\end{equation}
Assumption $1$ is expressed as
\begin{equation}
	\x^0_\The(\hat{\x}_t,t) = \gamma_t \x_0 + \phi_t \eps_{t}.
	\label{eq21:assumption}
\end{equation}
The predicted sample in the reverse process is
\begin{equation}
	\hat{\x}_{t-1} = \frac{\sqrt{\bar{\alpha}_{t - 1}}\beta_{t}}{1 - \bar{\alpha}_{t}}\x^0_\The(\x_t,t) + \frac{\sqrt{\alpha_{t}}(1 - \bar{\alpha}_{t - 1})}{1 - \bar{\alpha}_{t}}\x_{t} + \sqrt{\tilde{\beta}_t}\eps_{1},
	\label{eq22:bias_reverse}
\end{equation}
where $\x^0_\The(\x_t,t)=\frac{x_t-\sqrt{\bar{\alpha}_t}\eps_\The(\x_t,t)}{\sqrt{\bar{\alpha}_t}}$. 
We can substitute Eq.~\eqref{eq20:forward_one} and Eq.~\eqref{eq21:assumption} into Eq.~\eqref{eq22:bias_reverse} to obtain
\begin{equation}
	\begin{aligned}
		\hat{\x}_{t-1} &= \frac{\sqrt{\bar{\alpha}_{t - 1}}\beta_{t}}{1 - \bar{\alpha}_{t}}(\gamma_t\x_0 + \phi_t\eps_t) + \frac{\sqrt{\alpha_{t}}(1 - \bar{\alpha}_{t - 1})}{1 - \bar{\alpha}_{t}}(\sqrt{\bar{\alpha}_t}\x_0 + \sqrt{1 - \bar{\alpha}_t}\eps) \\
		&+ \sqrt{\tilde{\beta}_t}\eps_{1}\\ &= \frac{\sqrt{\bar{\alpha}_{t - 1}}\beta_{t}\gamma_t}{1 - \bar{\alpha}_{t}}\x_0   + \frac{\sqrt{\bar{\alpha}_{t - 1}}\beta_{t}\phi_t}{1 - \bar{\alpha}_{t}}\eps_t + \frac{\sqrt{\alpha_{t}}(1 - \bar{\alpha}_{t - 1})\sqrt{\bar{\alpha}_t}}{1 - \bar{\alpha}_{t}}\x_0\\& + \frac{\sqrt{\alpha_{t}}(1 - \bar{\alpha}_{t - 1})\sqrt{1 - \bar{\alpha}_t}}{1 - \bar{\alpha}_{t}}\eps + \sqrt{\tilde{\beta}_t}\eps_{1}\\ &= \left(\frac{\sqrt{\bar{\alpha}_{t - 1}}\beta_{t}\gamma_t}{1 - \bar{\alpha}_{t}} + \frac{\sqrt{\alpha_{t}}(1 - \bar{\alpha}_{t - 1})\sqrt{\bar{\alpha}_t}}{1 - \bar{\alpha}_{t}}\right)\x_0 + \frac{\sqrt{\bar{\alpha}_{t - 1}}\beta_{t}\phi_t}{1 - \bar{\alpha}_{t}}\eps_t \\& + \frac{\sqrt{\alpha_{t}}(1 - \bar{\alpha}_{t - 1})\sqrt{1 - \bar{\alpha}_t}}{1 - \bar{\alpha}_{t}}\eps + \sqrt{\tilde{\beta}_t}\eps_{1}.
		\label{eq23:tui}
	\end{aligned} 
\end{equation}

Now, we deal with the first part of Eq.~\eqref{eq23:tui}:
\begin{equation}
	\begin{aligned}
		&\frac{\sqrt{\bar{\alpha}_{t - 1}}\beta_{t}\gamma_t}{1 - \bar{\alpha}_{t}} + \frac{\sqrt{\alpha_{t}}(1 - \bar{\alpha}_{t - 1})\sqrt{\bar{\alpha}_t}}{1 - \bar{\alpha}_{t}}  = \frac{\sqrt{\bar{\alpha}_{t - 1}}\beta_{t}\gamma_t+\sqrt{\alpha_{t}}(1 - \bar{\alpha}_{t - 1})\sqrt{\bar{\alpha}_t}}{1 - \bar{\alpha}_{t}} \\
		&=\frac{\sqrt{\bar{\alpha}_{t - 1}}(1-{\alpha}_{t})\gamma_t+{\alpha_{t}}(1 - \bar{\alpha}_{t - 1})\sqrt{\bar{\alpha}_{t-1}}}{1 - \bar{\alpha}_{t}}\\
		&=\frac{\sqrt{\bar{\alpha}_{t - 1}}\big((1-{\alpha}_{t})\gamma_t+{\alpha_{t}}(1 - \bar{\alpha}_{t - 1})\big)}{1 - \bar{\alpha}_{t}}.
		\label{eq24}
	\end{aligned}
\end{equation}

Then, we need to construct an auxiliary term. Therefore, we discard $\gamma_t$ in Eq.~\eqref{eq24}:

\begin{equation}
	\begin{aligned}
		\frac{\sqrt{\bar{\alpha}_{t - 1}}\big((1-{\alpha}_{t})+{\alpha_{t}}(1 - \bar{\alpha}_{t - 1})\big)}{1 - \bar{\alpha}_{t}}
		&=\frac{\sqrt{\bar{\alpha}_{t - 1}}\big(1 - \alpha_{t}\bar{\alpha}_{t - 1}\big)}{1 - \bar{\alpha}_{t}} =\sqrt{\bar{\alpha}_{t - 1}}.
		\label{eq25}
	\end{aligned}
\end{equation}

Since $1-\alpha_{t}>0, \gamma_t\leq1$, according to Eq.~\eqref{eq24} and Eq.~\eqref{eq25}, we can naturally obtain

\begin{equation*}
	\begin{aligned}
		\frac{\sqrt{\bar{\alpha}_{t - 1}}\big((1-{\alpha}_{t})\gamma_t+{\alpha_{t}}(1 - \bar{\alpha}_{t - 1})\big)}{1 - \bar{\alpha}_{t}}\leq\sqrt{\bar{\alpha}_{t - 1}}
	\end{aligned}
\end{equation*}

We can definitely define a new coefficient $\gamma_{t-1}\leq1$ such that
\begin{equation}
	\begin{aligned}
		\frac{\sqrt{\bar{\alpha}_{t - 1}}\big((1-{\alpha}_{t})\gamma_t+{\alpha_{t}}(1 - \bar{\alpha}_{t - 1})\big)}{1 - \bar{\alpha}_{t}}=\gamma_{t-1}\sqrt{\bar{\alpha}_{t - 1}}.
	\end{aligned}
	\label{eq26}
\end{equation}

For the second part of Eq.~\eqref{eq23:tui}, we can always split it into

\begin{equation}
	\begin{aligned} Var(\hat{\x}_{t-1})&=(\frac{\sqrt{\bar{\alpha}_{t-1}}\beta_{t}}{1-\bar{\alpha}_{t}}\phi_{t})^{2}+(\frac{\sqrt{\alpha_{t}}(1-\bar{\alpha}_{t-1})}{1-\bar{\alpha}_{t}}\sqrt{1-\bar{\alpha}_{t-1}})^{2}+\tilde{\beta}_{t}\\ &=(\frac{\sqrt{\bar{\alpha}_{t-1}}\beta_{t}}{1-\bar{\alpha}_{t}}\phi_{t})^{2}+(\frac{\sqrt{\alpha_{t}}(1-\bar{\alpha}_{t-1})}{1-\bar{\alpha}_{t}}\sqrt{1-\bar{\alpha}_{t-1}})^{2}\\ &+\frac{(1-\bar{\alpha}_{t-1})(1-\alpha_{t})}{1-\bar{\alpha}_{t}}\\ &=(\frac{\sqrt{\bar{\alpha}_{t-1}}\beta_{t}}{1-\bar{\alpha}_{t}}\phi_{t})^{2}+\frac{\alpha_{t}(1-\bar{\alpha}_{t-1})^{2}}{1-\bar{\alpha}_{t}}+\frac{(1-\bar{\alpha}_{t-1})(1-\alpha_{t})}{1-\bar{\alpha}_{t}}\\ &=(\frac{\sqrt{\bar{\alpha}_{t-1}}\beta_{t}}{1-\bar{\alpha}_{t}}\phi_{t})^{2}+\frac{\alpha_{t}(1-\bar{\alpha}_{t-1})^{2}+(1-\bar{\alpha}_{t-1})(1-\alpha_{t})}{1-\bar{\alpha}_{t}}\\ &=(\frac{\sqrt{\bar{\alpha}_{t-1}}\beta_{t}}{1-\bar{\alpha}_{t}}\phi_{t})^{2}+\frac{(1-\bar{\alpha}_{t-1})[\alpha_{t}(1-\bar{\alpha}_{t-1})+(1-\alpha_{t})]}{1-\bar{\alpha}_{t}}\\ &=(\frac{\sqrt{\bar{\alpha}_{t-1}}\beta_{t}}{1-\bar{\alpha}_{t}}\phi_{t})^{2}+\frac{(1-\bar{\alpha}_{t-1})[\alpha_{t}-\bar{\alpha}_{t}+1-\alpha_{t}]}{1-\bar{\alpha}_{t}}\\ &=(\frac{\sqrt{\bar{\alpha}_{t-1}}\beta_{t}}{1-\bar{\alpha}_{t}}\phi_{t})^{2}+\frac{(1-\bar{\alpha}_{t-1})[1-\bar{\alpha}_{t}]}{1-\bar{\alpha}_{t}}\\ &=(\frac{\sqrt{\bar{\alpha}_{t-1}}\beta_{t}}{1-\bar{\alpha}_{t}}\phi_{t})^{2}+1-\bar{\alpha}_{t-1}.
		\label{eq27}
	\end{aligned}
\end{equation}
Based on Eqs.~\eqref{eq26} and ~\eqref{eq27}, we can obtain 
\begin{equation}
	\begin{aligned}
		\hat{\x}_{t-1} =\gamma_{t-1}\sqrt{\bar{\alpha}_{t - 1}}\x_0+\sqrt{1-\bar{\alpha}_{t-1}+(\frac{\sqrt{\bar{\alpha}_{t-1}}\beta_{t}}{1-\bar{\alpha}_{t}}\phi_{t})^{2}}\eps_{t-1}
	\end{aligned}
\end{equation}

\section*{Appendix B}
In this section, we present the parameter settings for the important experiments in this paper, as shown in Tables~\ref{tab10:adm_app}, \ref{tab11:edm_pfgm_app}, \ref{tab12:ed-ddpm_app}, \ref{tab13:ddim_amed_app}, and \ref{tab14:edm_pfgm_app}.

\begin{table}[H]
	\centering
	\caption{\textsc{Settings on CIFAR-10 and Image-Net using ADM.}}
	\begin{tabularx}{0.45\textwidth}{lXXXXXXXXX}  
		\toprule
		& $w_l/w_h$ & \multicolumn{3}{c}{CIFAR-10} & \multicolumn{3}{c}{Image-Net} \\
		\cmidrule(lr){3-5} \cmidrule(lr){6-8}
		$T'$ & & 20 & 30 & 50 & 20 & 30 & 50 \\
		\midrule
		ADM &$w_l$  & 1.013 & 1.008 & 1.0036 & 0.050 & 0.040 & 0.028 \\
		ADM &$w_h$  & 1.064 & 1.034 & 1.015 & 0.997 & 0.998 & 1.001 \\
		\bottomrule
	\end{tabularx}
	\label{tab10:adm_app}
\end{table}


\begin{table}[H]
	\centering
	\caption{\textsc{Settings on CIFAR-10 using DDPM and IDDPM.}}
	
	\begin{tabularx}{0.45\textwidth}{lXXXXX}  
		\toprule
		& $w_l/w_h$ & \multicolumn{2}{c}{DDPM} & \multicolumn{2}{c}{IDDPM} \\
		
		\cmidrule(lr){3-4} \cmidrule(lr){5-6}  
		$T'$ & & 10 & 20 & 30  & 100 \\  
		Baseline &$w_l$ & 1.068 & 1.019  & 1.009 & 1.0011 \\ 
		Baseline &$w_h$ & 1.250 & 1.140  & 1.042 & 1.0060 \\ 
		\bottomrule
	\end{tabularx}
	\label{tab11:edm_pfgm_app}
\end{table}

\begin{table}[H]
	\centering
	\caption{\textsc{Settings on CIFAR-10 using A-DPM, EA-DPM.}}
	\begin{tabular}{@{\hspace{0.62em}}l@{\hspace{0.62em}}c@{\hspace{0.62em}}c@{\hspace{0.62em}}c@{\hspace{0.62em}}c@{\hspace{0.62em}}c@{\hspace{0.62em}}c@{\hspace{0.62em}}c@{\hspace{0.62em}}}
		\toprule
		& $w_l/w_h$ & \multicolumn{3}{c}{CIFAR10 (LS)} & \multicolumn{3}{c}{CIFAR10 (CS)} \\
		\cmidrule(lr){3-5} \cmidrule(lr){6-8}
		$T'$ &  & 10 & 25 & 50 & 10 & 25 & 50 \\
		\midrule
		A-DPM & $w_l$ & 0.132& 0.049& 0.03& 0.072& 0.032& 0.008 \\
		A-DPM &$w_h$ & 1.11& 1.038& 1.019& 1.202& 1.046& 1.018\\
		\midrule
		NPR-DPM & $w_l$ & 0.132& 0.048& 0.024& 0.066& 0.03& 0.007\\
		NPR-DPM &$w_h$ & 1.105& 1.034& 1.015 & 1.192& 1.045& 1.017\\
		\midrule
		SN-DPM &$w_l$ & 0.109& 0.043& 0.025& 0.052& 0.027& 0.009\\
		SN-DPM &$w_h$ & 1.013& 1.005& 1.005& 1.101& 1.019& 1.01\\
		\bottomrule
	\end{tabular}
	\label{tab12:ed-ddpm_app}
\end{table}

\begin{table}[H]
	\centering
	\caption{\textsc{Settings on CIFAR-10 using DDIM and AMED.}}
	\begin{tabularx}{0.45\textwidth}{lXXXXXXX}  
		\toprule
		&$w_l/w_h$ & \multicolumn{3}{c}{DDIM} & \multicolumn{3}{c}{AMED} \\
		\cmidrule(lr){3-5} \cmidrule(lr){6-8}
		$T'$ & & 10 & 25 & 50 & 5 & 7 & 9 \\
		\midrule
		Baseline &$w_l$ & 0.0 & 0.0 & 0.0 & 0.0014 & 0.0012 & 0.0027 \\
		Baseline &$w_h$ & 1.21 & 1.037 & 1.011 & 0.9955 & 0.9932 & 0.9985 \\
		\bottomrule
	\end{tabularx}
	\label{tab13:ddim_amed_app}
\end{table}

\begin{table}[H]
	\centering
	\caption{\textsc{Settings on CIFAR-10 using EDM and PFGM++.}}
	\begin{tabularx}{0.45\textwidth}{lXXXXXXX}  
		\toprule
		& $w_l/w_h$ & \multicolumn{3}{c}{EDM} & \multicolumn{3}{c}{PFGM++} \\
		\cmidrule(lr){3-5} \cmidrule(lr){6-8}
		$T'$ & & 13 & 21 & 35 & 13 & 21 & 35 \\
		Baseline &$w_l$ & 0.036 & 0.016 & 0.007& 0.037& 0.016 & 0.006 \\
		Baseline &$w_h$ & 1.087 & 1.054& 1.022 & 1.095 & 1.057 & 1.025 \\
		\bottomrule
	\end{tabularx}
	\label{tab14:edm_pfgm_app}
\end{table}

\begin{table}[H]
	\centering
	\caption{The search experiment of $w_l$ on CIFAR-10(LS) using A-DPM with 25 steps.}
	\begin{tabular}{@{\extracolsep{\fill}}lccccccc@{}}  
		\toprule
		$w_l$ & 0.0 & 0.03 & 0.04 & 0.049 & 0.05 & 0.06 & 0.07 \\  
		\midrule
		Time & 11.60 & 9.00 & 8.59 & 8.46 & 8.47 & 8.59 & 8.99 \\        
		\bottomrule
	\end{tabular}
	\label{tab15:par}
\end{table}

\section*{Appendix C}
We emphasize that W++ can quickly search for the optimal parameters, and in this section, we provide more favorable evidence. Specifically, $w_t^l$ and $w_t^h$ are adaptively determined by the inverse variance, and and regulated by $w_l$, $w_h$, and $t_{mid}$, which are searched efficiently. In particular, we select CIFAR-10 as the test dataset and A-DPM~\cite{baoanalytic} as the baseline model, and then the 25-step sampling task will be performed.

(1) $w_l$ and $w_h$. Due to the method's insensitivity to parameters, the parameter search process is fast via the two-stage search. Firstly, a coarse search with a step size of $0.01$ was performed. After finding a turning point in FID near $0.05$, a fine search with a step size of $0.001$ was conducted, quickly determining the optimal value as 0.049, as shown in Table~\ref{tab15:par}.

(2) $t_{\rm mid}$. Based on observations, model focuses on generating low-frequency contours during the first $80\%$ of the sampling and refines high-frequency details in the remaining $20\%$, as shown Fig.~\ref{fig2:evolution_dpm} of paper. Based on test results, we uniformly set to $0.2$. TUWN~\cite{yi2024towards} suggests the final $25\%$ of the sampling is critical for generating high-frequency details, which aligns closely with our $20\%$

(3) $\sigma_t$. $\sigma_t$ is known and directly set to the empirical inverse variance of the baseline model.
\end{document}